\documentclass[11pt]{article}

\usepackage{acl}

\usepackage{times}
\usepackage{latexsym}
\usepackage[T1]{fontenc}
\usepackage[utf8]{inputenc}
\usepackage{microtype}
\usepackage{inconsolata}
\usepackage{graphicx}

\usepackage{amsmath}
\usepackage{amssymb}
\usepackage{amsfonts}
\usepackage{booktabs}
\usepackage{subcaption}
\usepackage{enumitem}
\usepackage{amsthm}

\newtheorem{proposition}{Proposition}
\newtheorem{lemma}[proposition]{Lemma}
\theoremstyle{remark}
\newtheorem*{remark}{Remark}

\title{Thinking Seeds: Restarting On-Policy RL from the Model's Own Past Reasoning States}

\author{
  \textbf{Lei Yang\textsuperscript{1}},
  \textbf{Wei Bi\textsuperscript{2}},
  \textbf{Chenxi Sun\textsuperscript{2}},
  \textbf{Renren Jin\textsuperscript{1}},
  \textbf{Deyi Xiong\textsuperscript{1}}
\\
  \textsuperscript{1}TJUNLP Lab, College of Intelligence and Computing, Tianjin University, Tianjin, China \\
  \textsuperscript{2}Kuaishou Technology
\\
  \small{
    \textbf{Correspondence:} \href{mailto:yanglei_9@tju.edu.cn}{yanglei\_9@tju.edu.cn}
  }
}

\begin{document}
\maketitle

\begin{abstract}
On-policy reinforcement learning (RL) for language models trains on the reasoning states visited by the current policy. As training sharpens the policy, some states reached by earlier checkpoints disappear from its rollouts. We ask whether those states can be brought back without treating the actions that created them as ordinary on-policy data. A token-level importance ratio corrects the selected action, but it does not correct the preceding context; the resulting state-distribution mismatch is bounded by the policy divergence accumulated along the sequence. We therefore use an earlier rollout as a starting state and let the current policy generate the continuation. We call this procedure \textbf{Thinking Seeds}. Its length-ratio variant improves over on-policy training and two off-policy extensions on mathematical reasoning with two backbones. On code generation, using the prefix only as a state reaches 52.19, compared with 50.20 when prefix actions are also trained and 48.58 for on-policy training. At the population level, the continuation has the policy-gradient form of a restart objective, while the treatment of prefix actions is an explicit design choice. Relay inference and token-position analyses provide evidence for the state-coverage and action-mismatch motivations.
\end{abstract}

\section{Introduction}

Reinforcement learning (RL) from human feedback~\citep{DBLP:conf/nips/ChristianoLBMLA17, DBLP:conf/nips/StiennonO0ZLVRA20, DBLP:conf/nips/Ouyang0JAWMZASR22, DBLP:journals/corr/abs-2204-05862} or verifiable rewards~\citep{DBLP:journals/corr/abs-2411-15124, DBLP:journals/corr/abs-2402-03300} is now central to the post-training of large language models (LLMs)~\citep{DBLP:conf/icml/LiXZL00L24, DBLP:conf/acl/AhmadianCGFKPUH24, DBLP:journals/corr/abs-2501-03262}. Group Relative Policy Optimization (GRPO)~\citep{DBLP:journals/corr/abs-2402-03300} and its variants are widely adopted because they remove the value network required by PPO~\citep{DBLP:journals/corr/abs-1707-06347} at competitive performance.

An on-policy method, however, learns only at the reasoning states its current policy visits. As training succeeds, the policy reinforces its existing behavior and suppresses low-probability samples~\citep{DBLP:journals/corr/abs-2504-07912, DBLP:journals/corr/abs-2504-13837,DBLP:journals/corr/abs-2503-01307}, so the set of partial solutions in its rollouts can narrow. In the relay experiment of Section~\ref{sec:motivation}, an earlier checkpoint writes the first half of a response and the final checkpoint completes it; at large $k$, this has higher pass@$k$ than letting the final checkpoint generate the whole response.

Recent work incorporates off-policy data to recover what on-policy sampling misses~\citep{DBLP:journals/corr/abs-2505-22257, DBLP:journals/corr/abs-2506-21495,DBLP:journals/corr/abs-2504-14945, DBLP:journals/corr/abs-2508-11016, DBLP:journals/corr/abs-2510-01161}. These methods use tokens from another policy as training actions: a token-level importance ratio corrects which token was selected, but the context leading to it was also generated by the other policy, and Proposition~\ref{prop:bias} bounds the remaining mismatch at a position by the divergence accumulated over preceding tokens. An off-policy trajectory can instead supply a starting state without its actions being trained.

\begin{figure*}[t]
    \centering
    \includegraphics[width=0.78\textwidth]{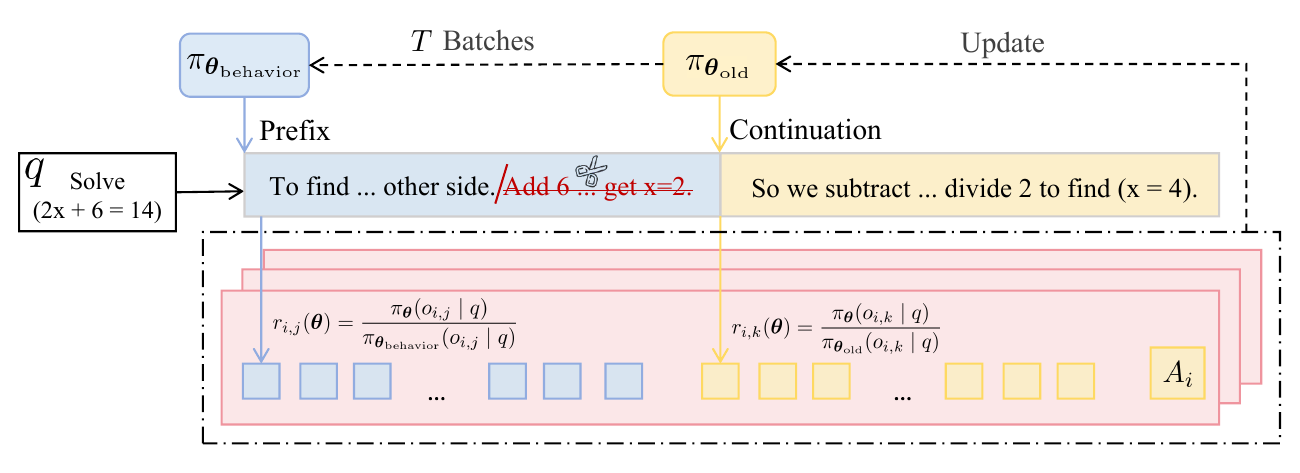}
    \caption{Overview of Thinking Seeds.}
    \label{fig:overview}
\end{figure*}

We propose \textbf{Thinking Seeds} (Figure~\ref{fig:overview}), which turns the model's recent rollouts into starting states. Every $T$ steps the current checkpoint generates rollouts for the next $T$ batches; the first batch is used directly, and for each later batch we truncate every response and let the policy current at training time continue from the kept prefix. In reinforcement-learning terms this replaces the prompt-only start distribution with a restart distribution~\citep{kakade2002approximately, agarwal2021theory}. The state${+}$action variant trains the prefix tokens through their behavior-policy ratios; the state variant masks them from the policy loss and keeps the prefix only as conditioning context.

Our contributions are:

\begin{enumerate}[wide=0pt,noitemsep,topsep=0pt]
    \item We separate the state and action roles of an off-policy trajectory: the token-level analysis identifies the context mismatch left by action-level reuse, while the restart formulation isolates the continuation generated by the current policy.
    \item We propose Thinking Seeds, which restarts from truncated rollouts of the model's own recent checkpoints and needs no external model. Its refresh interval $T$ and truncation ratio trade older, broader starting states against the amount and freshness of the continuation.
    \item On mathematical reasoning (Qwen2.5-Math-7B and DeepSeek-R1-Distill-Qwen-1.5B) the length-ratio variant improves over on-policy training and two off-policy baselines; on code generation (Qwen3-4B-Thinking-2507) the same hyperparameters transfer, and masking the prefix from the loss raises LiveCodeBench-v6 from $48.58$ to $52.19$. The wall-clock overhead is 10--20\% per step.
\end{enumerate}

\section{Related Work}

\paragraph{Off-policy Data as a Supplement.}
Recent work introduces off-policy samples into the GRPO framework to improve exploration and sample efficiency~\citep{DBLP:journals/corr/abs-2505-22257, DBLP:journals/corr/abs-2506-21495, DBLP:journals/corr/abs-2508-11016, DBLP:journals/corr/abs-2509-23232, DBLP:journals/corr/abs-2506-05433, DBLP:journals/corr/abs-2405-08448}. LUFFY~\citep{DBLP:journals/corr/abs-2504-14945} mixes in trajectories from external strong LLMs with regularized importance sampling, and M2PO~\citep{DBLP:journals/corr/abs-2510-01161} constrains the second moment of the ratios. Others inject partial thinking steps or prefixes from stronger models as hints~\citep{DBLP:journals/corr/abs-2507-02841, DBLP:journals/corr/abs-2509-06923, zhang2025adhint, huang2025blending, chen2025lppo}, or reweight expert and SFT tokens at the loss level~\citep{zhang2025chord, wu2025dft}, without controlling \emph{where} off-policy influence occurs in the sequence. Closest to our setting, Prefix-RFT~\citep{huang2025blending} also completes off-policy prefixes on-policy, but its prefixes come from a fixed set of external demonstrations and carry a supervised loss, whereas ours come from the learner's own checkpoints and are refreshed online.

\paragraph{Partial Rollout.}
Asynchronous RL for long chain-of-thought or agentic rollouts~\citep{DBLP:journals/corr/abs-2505-24298, slime_github, DBLP:conf/iclr/NoukhovitchHXHA25, DBLP:journals/corr/abs-2504-15930, DBLP:journals/corr/abs-2508-18588} introduces implicit off-policy data by generating single trajectories across several checkpoints, and mitigates the effect with decoupled PPO objectives~\citep{DBLP:journals/corr/abs-2505-24298}. Thinking Seeds is the \emph{proactive} counterpart: it deliberately places historical tokens in the prefix and keeps the rest on-policy.

\paragraph{Restart Distributions.}
In classical RL the start distribution used during optimization governs what policy gradient methods can learn: conservative policy iteration optimizes from a restart distribution~\citep{kakade2002approximately}, and convergence to a global optimum depends on how well that distribution covers the states visited by good policies~\citep{agarwal2021theory}. In LLM RL every partial response is a state, so a truncated rollout is a valid restart state drawn from the learner's own recent behavior.

\section{Motivation}
\label{sec:motivation}

Two experiments separate what an off-policy trajectory offers: the states it passes through help the current policy, while reusing its actions is corrected least accurately exactly where the policy still has a choice to make.

\subsection{Earlier Checkpoints Reach States the Current Policy No Longer Visits}

\begin{figure}[t]
    \centering
    \includegraphics[width=0.8\columnwidth]{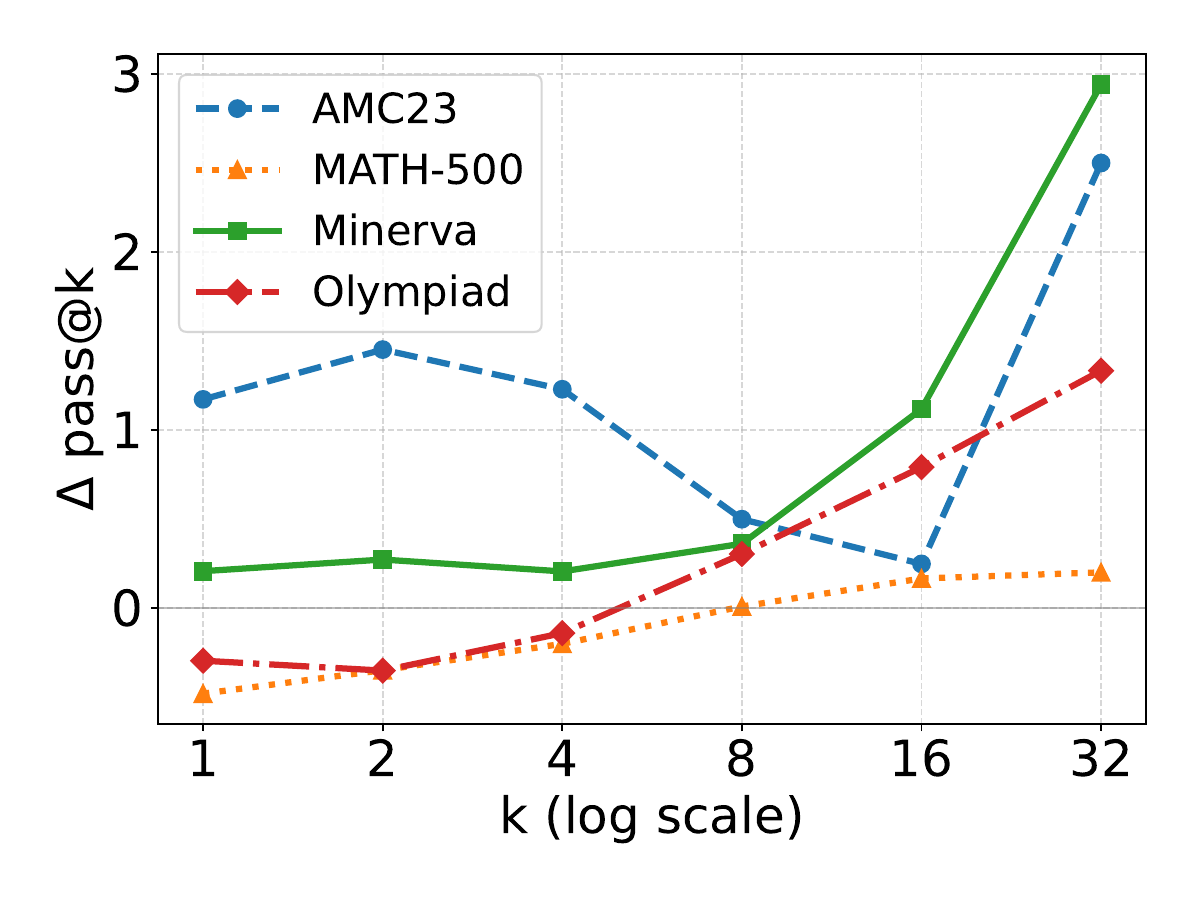}
    \caption{Difference in pass@$k$ between the two inference strategies. Positive values favor relay inference.}
    \label{fig:single_multi}
\end{figure}

On-policy RL sharpens the policy as training progresses: entropy decreases~\citep{DBLP:journals/corr/abs-2504-07912, DBLP:journals/corr/abs-2504-13837} and the partial solutions it generates concentrate on fewer paths. A state --- a prompt with a partial response --- that the policy no longer reaches contributes no gradient, however useful it would be.

We compare two inference strategies with a trained model: \textbf{single-model inference}, where the final checkpoint generates all tokens, and \textbf{relay inference}, where a strong earlier checkpoint generates the first half of the response and the final checkpoint continues from that state, so that only the distribution of intermediate states changes.

Figure~\ref{fig:single_multi} shows the pass@$k$ difference (detailed scores in Appendix~\ref{sec:det_comp}). At $k<8$ neither strategy has a consistent advantage; at $k\geq8$ relay inference is consistently better on the challenging benchmarks, so prefixes reached by an earlier checkpoint can lead the final policy to solutions it rarely finds from the prompt alone. The training implication is a hypothesis rather than a consequence of this inference experiment: if independently seeded rollouts raise the chance of a correct response, they may also reduce all-wrong groups, a relationship Lemma~\ref{lem:group} makes explicit.

\subsection{Reusing Off-Policy Actions Is Costly}
\label{sec:entropy_collapse}

\begin{figure}[t]
    \centering
    \begin{subfigure}[b]{0.72\columnwidth}
        \centering
        \includegraphics[width=\linewidth]{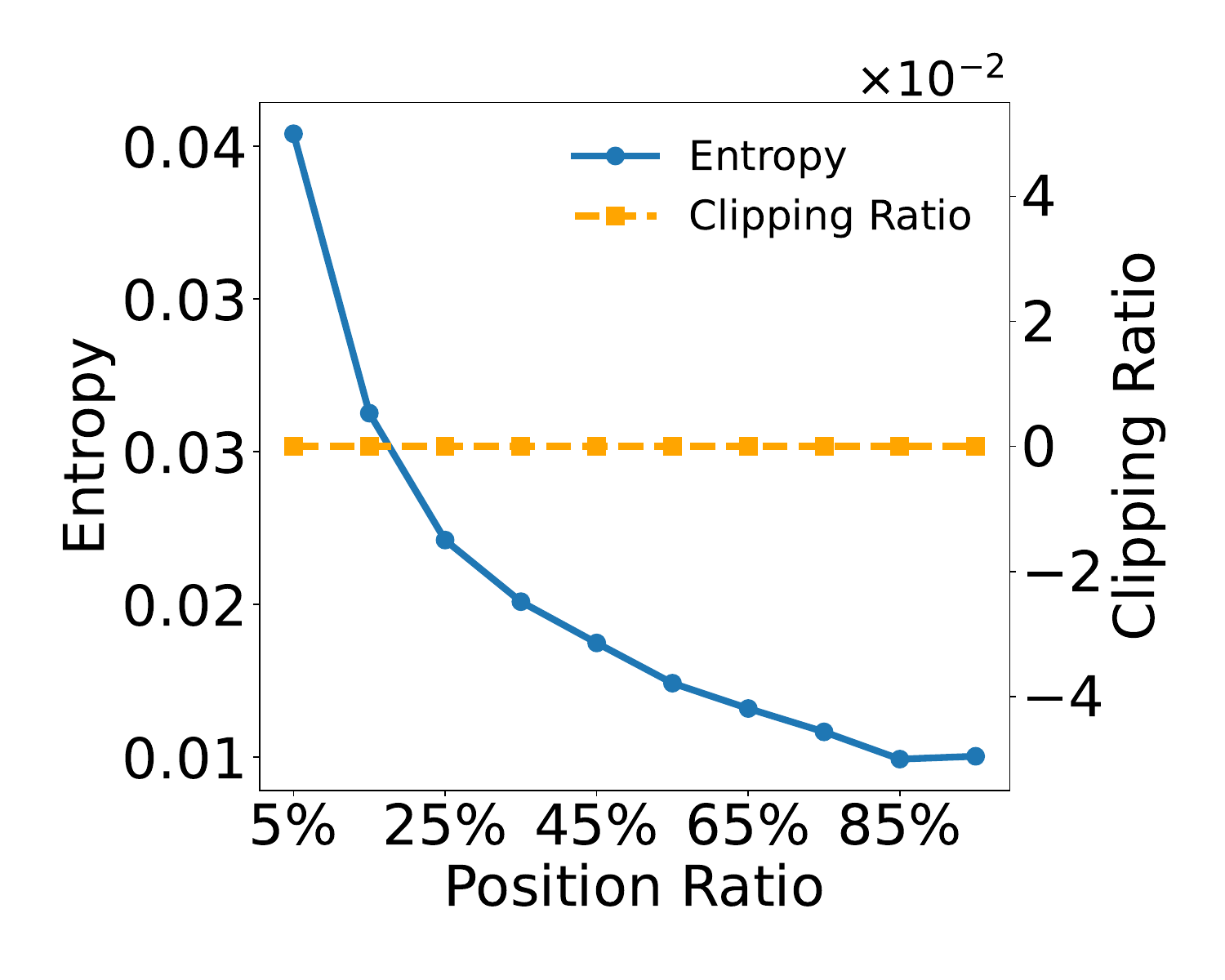}
        \caption{On-policy w/o Mini Batch}
        \label{fig:on_wo_mini}
    \end{subfigure}

    \vspace{0.2em}
    \begin{subfigure}[b]{0.72\columnwidth}
        \centering
        \includegraphics[width=\linewidth]{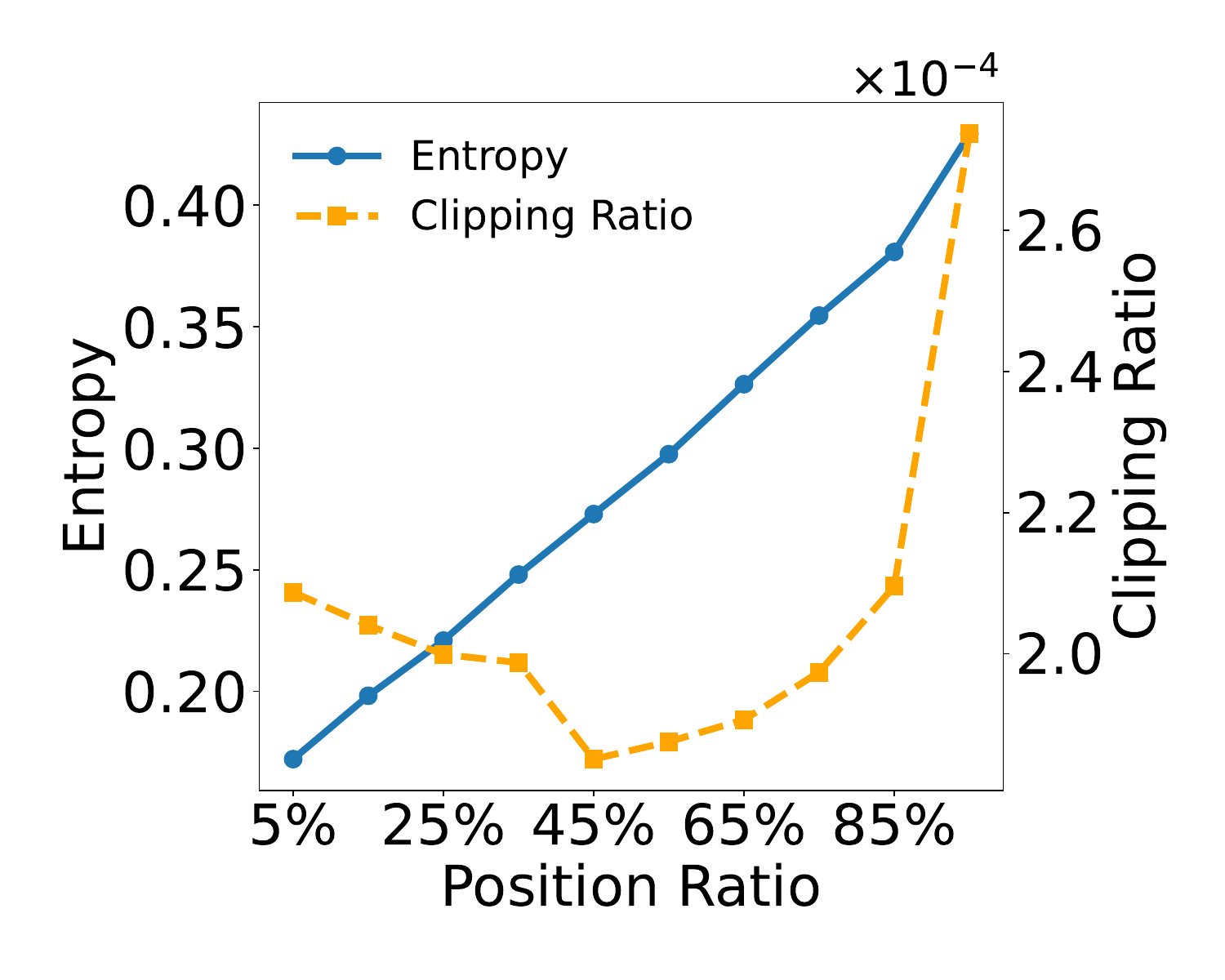}
        \caption{On-policy w/ Mini-batch}
        \label{fig:on_w_mini}
    \end{subfigure}
    \caption{Entropy and clipping ratio vs.\ relative token position (10\% bins) under on-policy training.}
    \label{fig:ana_en_mi}
\end{figure}

Borrowing an action taken by another policy requires correcting for the distribution it was drawn from. Practical methods keep token-level ratios and control the update magnitude by constraining their second moment~\citep{DBLP:journals/corr/abs-2510-01161} or regularizing them~\citep{DBLP:journals/corr/abs-2504-14945}; they do not change the state distribution from which a token is evaluated.

We measure average entropy and the proportion of clipped tokens per 10\% bin of relative token position (Figure~\ref{fig:ana_en_mi}), in two on-policy settings: one without mini-batching (batch size 32) and one with mini-batching (mini-batch size 32, batch size 512).

Without mini-batching (Figure~\ref{fig:on_wo_mini}), every token is generated and trained on-policy, and entropy decreases along positions: the policy becomes increasingly deterministic as generation proceeds.

With mini-batching (Figure~\ref{fig:on_w_mini}), each batch is consumed by several successive updates, so after the first update the later ones reuse actions of a drifted policy --- a controlled example of stale action reuse. Entropy now rises along positions instead of falling, and the clipping ratio rises with it, from about $1.8$ to $2.8$ tokens in $10^4$. Clipping is therefore concentrated at late, high-entropy positions, consistent with the position dependence in Proposition~\ref{prop:bias} and with prior reports of the same coincidence~\citep{DBLP:journals/corr/abs-2510-01161}. This motivates protecting continuation decisions from stale action reuse.

\subsection{Design Principle: Borrow States, Not Actions}

An off-policy trajectory offers both the states it passes through and the actions chosen at them. The first experiment suggests that earlier states can help the final policy; the second shows why older actions require care. Thinking Seeds therefore keeps a prefix as a starting state and samples the continuation from the current policy, with the prefix actions either entering the loss through their behavior-policy ratios (the state${+}$action variant) or only conditioning the continuation (the state variant).

\section{Methodology}

We first fix the on-policy objective Thinking Seeds builds on, then describe how seeds are drawn, how their tokens enter the loss, and where they are cut, and finally analyze what the construction changes relative to on-policy training.

\subsection{Preliminaries}
\label{sec:prelim}

GRPO~\citep{DBLP:journals/corr/abs-2402-03300} estimates advantages group-relatively, removing the need for a separate value network. We build on its DAPO variant~\citep{DBLP:journals/corr/abs-2503-14476}:

\begin{equation}
\label{equ_dapo}
\begin{aligned}
& \mathcal{J}_{\text{DAPO}}({\boldsymbol{\theta}})
= \mathbb{E}_{(q,a)\sim P(Q),\, \{o_i\}_{i=1}^G \sim \pi_{{\boldsymbol{\theta}}_{\text{old}}}} \\
& \left[
\frac{1}{\sum_{i=1}^G |o_i|}
\sum_{i=1}^G
\sum_{t=1}^{|o_i|}
\min\!\Big(
    r_{i,t}({\boldsymbol{\theta}})\,\hat{A}_{i,t},
\right. \\
& \left.
\operatorname{clip}\!\big(
        r_{i,t}({\boldsymbol{\theta}}),\, 1{-}\varepsilon_{\text{low}},\, 1{+}\varepsilon_{\text{high}}
    \big)\hat{A}_{i,t}
\Big)
\right]
\end{aligned}
\end{equation}

\noindent
where $R_i$ is the overall reward for response $o_i$ and the advantage estimate is $\hat{A}_{i,t}=(R_i-\text{mean}(\{R_i\}_{i=1}^G))/\text{std}(\{R_i\}_{i=1}^G)$. We write ${\boldsymbol{\theta}}$ for the parameters being optimized, ${\boldsymbol{\theta}}_{\text{old}}$ for the snapshot of ${\boldsymbol{\theta}}$ that sampled the current batch and supplies the denominator of the importance ratio $r_{i,t}({\boldsymbol{\theta}}) = \pi_{\boldsymbol{\theta}}(o_{i,t} \mid q, o_{i,<t}) / \pi_{{\boldsymbol{\theta}}_{\text{old}}}(o_{i,t} \mid q, o_{i,<t})$, and ${\boldsymbol{\theta}}_{\text{behavior}}$ for the parameters of any policy that generated data used in training. ${\boldsymbol{\theta}}_{\text{old}}$ deviates from ${\boldsymbol{\theta}}$ only when one batch is consumed by several mini-batch updates; in standard on-policy training ${\boldsymbol{\theta}}_{\text{behavior}} \equiv {\boldsymbol{\theta}}_{\text{old}}$, and clipping keeps $r_{i,t}$ within a trust region.

When the behavior policy differs from $\pi_{{\boldsymbol{\theta}}_{\text{old}}}$, Equation~\ref{equ_dapo} is still applied formally, with $\pi_{{\boldsymbol{\theta}}_{\text{behavior}}}$ in the denominator, but the data no longer come from the policy whose gradient is being estimated. This mismatch can destabilize training~\citep{DBLP:conf/icml/SchulmanLAJM15}, which is why several studies favor purely on-policy training~\citep{DBLP:journals/corr/abs-2505-23585, DBLP:journals/corr/abs-2503-20783}, while pure on-policy RL tends to reinforce existing behavior and under-explore~\citep{DBLP:journals/corr/abs-2504-07912}.

\paragraph{States and Actions.}
We view generation as a sequential decision process. The policy factorizes autoregressively,
\begin{equation}
\pi(o \mid q) = \prod_{t=1}^{|o|} \pi(o_t \mid s_t), \quad s_t = (q, o_{<t}),
\end{equation}
so the state $s_t$ is the prompt together with the partial response and the action is the next token $o_t$. Transitions are deterministic and the reward $R(q,o)$ is given at the end. We write $d^{\pi}_t$ for the distribution of $s_t$ when generation starts from the prompt and follows $\pi$. A token-level importance ratio corrects the choice of $o_t$ at $s_t$, but not the fact that $s_t$ was drawn from $d^{\pi_{\text{behavior}}}_t$ rather than $d^{\pi_{\boldsymbol{\theta}}}_t$.

\subsection{Thinking Seeds: Restarting from Past Reasoning States}

Thinking Seeds draws starting states from the model's own recent rollouts and runs ordinary on-policy RL from them. Training proceeds in cycles of $T$ steps. At the start of a cycle the current policy becomes the behavior policy, written $\pi_b$ for $\pi_{{\boldsymbol{\theta}}_\text{behavior}}$, and generates rollouts for the next $T$ batches in a single generation phase. The first batch is trained as it is, which is ordinary on-policy training. For each of batches $2,\dots,T$, every behavior response is truncated and the kept prefix serves as the starting state from which the policy current at that step continues:
\begin{equation}
\label{eq:s_behav}
\begin{gathered}
\tilde o \sim \pi_b(\cdot \mid q), \quad h \sim \operatorname{truncate}(\tilde o), \\
y \sim \pi_{{\boldsymbol{\theta}}_{\text{old}}}(\cdot \mid q, h),
\end{gathered}
\end{equation}
and the training sample is $o=[h, y]$. We call $h$ a \emph{seed} and write $\rho_b(h \mid q)$ for the seed distribution induced by $\pi_b$ and the truncation rule; it does not depend on ${\boldsymbol{\theta}}$. Each of the $G$ rollouts of a prompt is grown from its own seed, so a group starts from $G$ different states. In our experiments the mini-batch size equals the training batch size, so ${\boldsymbol{\theta}}_{\text{old}}\equiv{\boldsymbol{\theta}}$ and the continuation $y$ is exactly on-policy. A seed in batch $k$ was written by a policy $k-1$ steps old, so $T$ bounds seed staleness.

Each token is assigned to the policy that generated it. For the $t$-th token of sample $o_i=[h_i,y_i]$, the importance ratio is
\begin{equation}
\begin{gathered}
r_{i,t}({\boldsymbol{\theta}})=
\begin{cases}
\displaystyle \frac{\pi_{\boldsymbol{\theta}}(o_{i,t} \mid q, o_{i,<t})}{\pi_b(o_{i,t} \mid q, o_{i,<t})}, & t \le |h_i|, \\[2ex]
\displaystyle \frac{\pi_{\boldsymbol{\theta}}(o_{i,t} \mid q, o_{i,<t})}{\pi_{{\boldsymbol{\theta}}_{\text{old}}}(o_{i,t} \mid q, o_{i,<t})}, & t > |h_i|,
\end{cases}
\end{gathered}
\end{equation}
and the samples are trained with the DAPO objective (Equation~\ref{equ_dapo}) and the same group-relative advantage. Conditional on the supplied seed, continuation tokens are on-policy. The seed tokens are a separate design choice: the default state${+}$action variant trains them through their ratios to $\pi_b$, while the state variant uses them only to provide the conditioning state. We express the choice with a token mask:
\begin{equation}
\label{eq:mask}
 m_{i,t}=\begin{cases}
 1, & \text{state}{+}\text{action},\\[0.4ex]
 \mathbf{1}\{t>|h_i|\}, & \text{state}.
 \end{cases}
\end{equation}
The mask multiplies the token sum in Equation~\ref{equ_dapo}, and the token mean is normalized by the number of unmasked tokens, $\sum_i\sum_t m_{i,t}$, so for the state variant the denominator is $\sum_i|y_i|$; this keeps the weight of a trained continuation token comparable to that in an on-policy batch.

\subsection{Truncation Strategy}

The truncation rule $\operatorname{truncate}(\tilde o)$ in Equation~\ref{eq:s_behav} determines the seed distribution $\rho_b$: how deep the starting states are and at which kinds of tokens the current policy takes over. Thinking Seeds does not depend on a specific rule; we consider two. \textbf{Length-ratio-based truncation} keeps the first $\operatorname{round}(|\tilde o| \times ratio)$ tokens of the behavior response as the seed, so $ratio$ sets both the depth of the starting states and the split between seed and continuation tokens. \textbf{Entropy-based truncation} starts from the observation that high-entropy tokens are important for learning~\citep{DBLP:journals/corr/abs-2505-22617, DBLP:journals/corr/abs-2505-22312} but are often clipped when reused as off-policy actions (Section~\ref{sec:entropy_collapse}); we select the $k$ highest-entropy positions of the behavior response and sample one uniformly as the truncation point, so that the current policy makes the decision at that branching point on-policy. We compare the two strategies, written Ours$_{\text{ratio}}$ and Ours$_{\text{entropy}}$, on both mathematical backbones.

\begin{table*}[t]
    \centering
    \footnotesize
    \begin{tabular*}{\textwidth}{@{\extracolsep{\fill}}lccccccc@{}}
    \toprule
        \textbf{Method} & \textbf{AIME24} & \textbf{AIME25} & \textbf{MATH-500} & \textbf{AMC23} & \textbf{Minerva} & \textbf{Olympiad} & \textbf{Avg} \\
        \midrule
        \multicolumn{8}{c}{\textbf{Qwen2.5-Math-7B}} \\
        \midrule
        On-policy   & 34.58 & 17.19 & \textbf{83.33} & 71.25 & 37.48 & 45.97 & 48.30 \\
        LUFFY       & 24.27 & 17.81 & 83.48 & 67.50 & 34.54 & \textbf{49.99} & 46.26 \\
        M2PO        & 35.83 & 17.60 & 79.23 & 73.75 & 36.89 & 43.76 & 47.84 \\
        Ours$_{\text{ratio}}$
                    & \textbf{36.35} & \textbf{19.90} & 83.25 & \textbf{79.84} & \textbf{39.28} & 48.32 & \textbf{51.16} \\
        Ours$_{\text{entropy}}$
                    & \textbf{36.35} & 17.92 & 82.89 & 75.39 & 36.82 & 46.14 & 49.25 \\
        \midrule
        \multicolumn{8}{c}{\textbf{DeepSeek-R1-Distill-Qwen-1.5B}} \\
        \midrule
        On-policy&35.94 & 27.92 & 88.08 & 78.83 & 33.86 & 58.00 & 53.77  \\
        LUFFY  & 27.50 & 26.25 & 82.38 & 72.27 & 31.24 & 51.61 & 48.54    \\
        M2PO  & 36.25 &	27.81 &	87.28 &	77.66 &	32.19 &	56.92 &	53.02    \\
        Ours$_{\text{ratio}}$ & \textbf{37.19} & \textbf{28.54} & \textbf{88.25} & \textbf{79.22} & \textbf{34.46} & \textbf{59.05} & \textbf{54.45}    \\
        Ours$_{\text{entropy}}$  & 31.35 & 26.04 & 83.59 & 71.48 & 32.07 & 52.20 & 49.46  \\
        \bottomrule
    \end{tabular*}
    \caption{Overall performance on six mathematical reasoning benchmarks and two backbones.}
    \label{tab:main_results}
\end{table*}

\subsection{Analysis: Borrowing States versus Borrowing Actions}
\label{sec:theory}

We now make this separation precise, with rewards $R \in [0,1]$, writing $\pi$ for $\pi_{\boldsymbol{\theta}}$ and $\nabla$ for $\nabla_{\boldsymbol{\theta}}$. Each batch is consumed by a single update, so on-policy ratios equal $1$ and lie inside the clipping range: the gradient of Equation~\ref{equ_dapo} on those tokens is the unclipped policy gradient. Proofs are in Appendix~\ref{sec:proofs}.

\paragraph{The restart objective.}
Seeding defines the objective
\begin{equation}
\mathcal{J}_{\rho}({\boldsymbol{\theta}}) = \mathbb{E}_{q,\,h\sim\rho_b,\,y\sim\pi(\cdot\mid q,h)} \big[R(q,[h,y])\big],
\end{equation}
the expected reward when generation restarts from a seed. With the empty seed it reduces to the usual objective $\mathcal{J}$.

\begin{proposition}[Continuation gradients under the restart objective]
\label{prop:exact}
For any baseline $b(q)$ that does not depend on $(h,y)$,
\begin{equation}
\begin{aligned}
\nabla \mathcal{J}_{\rho} = \mathbb{E}_{q,h,y}\Big[&\big(R - b(q)\big)\\[-0.3ex]
&\times\!\!\sum_{t>|h|} \nabla \log \pi(o_t \mid s_t)\Big].
\end{aligned}
\end{equation}
\end{proposition}

\noindent
The implemented continuation-token update is the finite-sample counterpart of this population gradient, with the same group baseline, standard-deviation scaling, clipping, and token-mean normalization that DAPO applies to on-policy rollouts; under the state variant of Equation~\ref{eq:mask} it is the only policy-loss component. The behavior policy does not enter through an importance ratio, so the context-mismatch term caused by reusing prefix actions is absent from the continuation loss; the coverage of the restart distribution still determines whether the new objective is useful.

\begin{proposition}[The bias of borrowed actions grows with position]
\label{prop:bias}
Let $o\sim\mu(\cdot\mid q)$ for a behavior policy $\mu$. For position $t$, let $g_t(s) = \mathbb{E}_{a\sim\pi(\cdot\mid s)}[\Psi_t(s,a)\nabla\log\pi(a\mid s)]$ for a bounded signal $\Psi_t$, and $G_t=\sup_s\|g_t(s)\|$. The token-level importance-weighted gradient at $t$ has mean $\mathbb{E}_{s\sim d^{\mu}_t}[g_t(s)]$ and the on-policy gradient is $\mathbb{E}_{s\sim d^{\pi}_t}[g_t(s)]$; their difference is at most
\begin{equation}
2G_t\,\mathrm{TV}(d^{\mu}_t, d^{\pi}_t)
\le G_t \Big(2\sum_{j<t} \bar K_j\Big)^{1/2},
\end{equation}
where $\bar K_j = \mathbb{E}_{s\sim d^{\mu}_j}\mathrm{KL}\big(\mu(\cdot\mid s)\,\|\,\pi(\cdot\mid s)\big)$, and $\mathrm{TV}(d^{\mu}_t, d^{\pi}_t)$ is nondecreasing in $t$.
\end{proposition}

\noindent
A sequence-level product $\prod_{j<t}\pi/\mu$ corrects the preceding actions at the cost of a second moment that grows rapidly with $t$~\citep{liu2018breaking}. The remaining state-distribution term is small near the start of a trajectory and can be larger later; mini-batch updates (Figure~\ref{fig:on_w_mini}) are an example with $\mu=\pi_{{\boldsymbol{\theta}}_\text{old}}$. In the state${+}$action variant the prefix-action contribution is confined to seed positions, $t\leq |h|$, and depends on $ratio$ and on the staleness of $\pi_b$, at most $T-1$ steps.

\paragraph{Comparing the two loss treatments.}
The state variant removes the prefix's direct policy-loss contribution and recomputes the token mean over the continuation tokens, leaving everything else unchanged.

\begin{proposition}[Seeds change where advantages are evaluated]
\label{prop:coverage}
Let $A^{\pi}(s,a)$ be the advantage of $\pi$, and let $d^{\pi}_{\rho,t}$ be the distribution of $s_t$, restricted to $t>|h|$, when generation starts at $h\sim\rho_b$ and continues with $\pi$. Then
\begin{equation}
\begin{aligned}
\nabla \mathcal{J}_{\rho} = \sum_{t} \mathbb{E}_{s\sim d^{\pi}_{\rho,t},\; a\sim\pi(\cdot\mid s)}\\
\big[A^{\pi}(s,a)\,\nabla\log\pi(a\mid s)\big].
\end{aligned}
\end{equation}
Over a refresh cycle with ${\boldsymbol{\theta}}$ held fixed, the expected continuation-token update is proportional to the gradient of $\mathcal{J}_T = \frac{1}{T}\mathcal{J} + \frac{T-1}{T}\mathcal{J}_{\rho}$, whose state weighting is $d_{T,t} = \frac{1}{T} d^{\pi}_{t} + \frac{T-1}{T} d^{\pi}_{\rho,t}$.
\end{proposition}

\noindent
Because a state contains the whole history, $A^{\pi}(s,a)$ does not depend on how $s$ was reached. Seeding therefore leaves the advantage function unchanged and only moves the distribution of states at which it is estimated, from $d^{\pi}_t$ to $d_{T,t}$.

\begin{remark}[The coverage trade-off]
On-policy training ($T{=}1$) assigns no continuation updates to states the current policy no longer reaches. A larger $T$ draws seeds from an older policy and can introduce such states, but leaves less weight on states reached only by the current policy; a larger $ratio$ gives a deeper starting state but fewer tokens for the current policy to generate, and in the state${+}$action variant exposes more of the loss to the mismatch in Proposition~\ref{prop:bias}. The trade-off is qualitative and does not determine an optimum for a particular LLM.
\end{remark}

\begin{lemma}[Zero-advantage groups]
\label{lem:group}
Let the reward be binary and the $G$ rollouts of a prompt be grown from independent seeds $h_i\sim\rho_b$, as in Thinking Seeds. With $p_\rho(q)=\mathbb{E}_{h\sim\rho_b}\Pr_{y\sim\pi}(R{=}1\mid q,h)$, the rewards are i.i.d.\ Bernoulli$(p_\rho)$ and the whole group has zero advantage with probability $D(p_\rho)=p_\rho^G+(1-p_\rho)^G$, where $(1-p_\rho)^G = 1-\text{pass@}G$ of seeded sampling. $D$ decreases on $[0,\frac12]$.
\end{lemma}

\noindent
For prompts the current policy rarely solves, a seed distribution with higher pass@$G$ therefore has fewer all-wrong groups at the same group size. This is a possible training benefit of the earlier-checkpoint states in Figure~\ref{fig:single_multi}.

\section{Experiments}

We evaluate Thinking Seeds on mathematical reasoning with two backbones and on code generation, and then ablate the two parameters that define the seed distribution, the truncation ratio and the refresh interval.

\subsection{Setup}
\label{sec:setup}

\paragraph{Models and Training.}
We used Qwen2.5-Math-7B~\citep{DBLP:journals/corr/abs-2409-12122} (4k context window, trained on DAPO-Math-17k~\citep{DBLP:journals/corr/abs-2503-14476} with each prompt--response sequence capped at 4k tokens) and DeepSeek-R1-Distill-Qwen-1.5B~\citep{DBLP:journals/corr/abs-2501-12948} (4,608 entries randomly sampled from Openr1-Math-46k-8192~\citep{DBLP:journals/corr/abs-2504-14945}, response length 8k).
Unless otherwise noted, we used a batch size of 32, a clipping-higher of 0.28, 8 responses per prompt, and $k=32$ for entropy-based truncation.
The main table uses the best $ratio$ and $T$ among the configurations tested per backbone (Section~\ref{sec:ablation_study}); remaining hyperparameters are in Appendix~\ref{sec:trainingpara}.

We additionally trained Qwen3-4B-Thinking-2507~\citep{yang2025qwen3} on code generation, with rewards computed by executing unit tests, on 5,120 prompts randomly sampled from KlearReasoner-CodeSub-15k~\citep{su2025klear} with maximum prompt and response lengths of 2k and 16k tokens. Thinking Seeds reuses $ratio=30\%$ and $T=8$ from the mathematical settings without further search, and the on-policy baseline uses the same configuration.

\paragraph{Compared Methods.}
Besides standard on-policy training, we compared against two methods that use off-policy samples directly, LUFFY~\citep{DBLP:journals/corr/abs-2504-14945} and M2PO~\citep{DBLP:journals/corr/abs-2510-01161}, with the same parameter configurations. For LUFFY we follow its data processing and filter out prompts whose off-policy data exceed the backbone context window.

\paragraph{Evaluation.}
We trained until convergence on the validation set, selecting the checkpoint with the highest combined average on AIME24 and AIME25~\citep{aops_aime}, and evaluated all models additionally on MATH-500~\citep{DBLP:conf/nips/HendrycksBKABTS21}, AMC23~\citep{aops_amc}, Minerva Math~\citep{DBLP:conf/nips/LewkowyczADDMRS22}, OlympiadBench~\citep{DBLP:conf/acl/HeLBHTSHHHZLQL024}, and, for code generation, LiveCodeBench-v6~\citep{jain2025livecodebench}. Every reported score is avg@32, the mean pass@1 over 32 samples per problem. Unless stated otherwise, the ablations and analyses use Qwen2.5-Math-7B.

\subsection{Main Results}
\label{sec:main_results}

Table~\ref{tab:main_results} compares the length-ratio variant with the baselines. It reaches 51.16 on Qwen2.5-Math-7B, compared with 48.30 for on-policy training, 46.26 for LUFFY, and 47.84 for M2PO, and 54.45 on DeepSeek-R1-Distill-Qwen-1.5B, compared with 53.77 for on-policy training. The entropy-based rule behaves differently across backbones: 49.25 on Qwen2.5-Math-7B and 49.46 on DeepSeek-R1-Distill-Qwen-1.5B, where it is below on-policy training. Section~\ref{sec:truncation_analysis} examines the truncation points the two rules select.

\begin{table}[t]
    \centering
    \small
    \begin{tabular}{lc}
    \toprule
        \textbf{Method} & \textbf{LiveCodeBench-v6} \\
        \midrule
        On-policy & 48.58 \\
        Ours$_{\text{ratio}}$, state${+}$action & 50.20 \\
        Ours$_{\text{ratio}}$, state & \textbf{52.19} \\
        \bottomrule
    \end{tabular}
    \caption{Code generation results on Qwen3-4B-Thinking-2507.}
    \label{tab:code_results}
\end{table}

\paragraph{Code Generation.}
Table~\ref{tab:code_results} reports the code generation setting, carrying over the mathematical configuration without further search. The state${+}$action variant improves LiveCodeBench-v6 from 48.58 to 50.20, and keeping the same prefixes as conditioning states while masking their direct loss contribution gives 52.19. The two seeded variants use the same rollouts, rewards, advantages, and generated tokens, so their 1.99-point difference compares the two loss treatments of Equation~\ref{eq:mask}, including the different token-mean normalization. In the terms of Section~\ref{sec:theory}, the state variant keeps only the restart-objective gradient of Proposition~\ref{prop:exact}, whereas state${+}$action also trains the prefix-action term whose bias Proposition~\ref{prop:bias} bounds.

\subsection{Ablation Study}
\label{sec:ablation_study}

\begin{table}[t]
    \centering
    \footnotesize
    \setlength{\tabcolsep}{4pt}
    \begin{tabular}{lccccc}
    \toprule
        $ratio$ & 0\% & 30\% & 50\% & 70\% & 90\% \\
        Avg & 48.30 & 50.56 & \textbf{50.81} & 50.73 & 49.67 \\
        \midrule
        $T$ & 1 & 2 & 4 & 8 & 16 \\
        Avg & 48.30 & 50.64 & 50.45 & \textbf{51.16} & 50.81 \\
        \bottomrule
    \end{tabular}
    \caption{Average over the six mathematical benchmarks for the two ablations (Qwen2.5-Math-7B); per-benchmark scores in Appendix~\ref{sec:ablation_details}.}
    \label{tab:ablation_avg}
\end{table}


\paragraph{Truncation Ratio.}
The truncation ratio sets how deep the starting states are and how many tokens of a seeded sample are trained on-policy. We varied it in the state${+}$action variant at $T=16$ with all other hyperparameters fixed. Table~\ref{tab:ablation_avg} shows performance is similar from 30\% to 70\% and drops at 90\% ($ratio=0\%$ is on-policy training), consistent with the qualitative trade-off in Section~\ref{sec:theory}: deeper seeds change the starting states, but leave fewer continuation tokens and expose more seed actions to behavior-policy mismatch.

\paragraph{Refresh Interval.}
The refresh interval $T$ sets how old the seeds can be (up to $T-1$ steps) and the fraction $(T-1)/T$ of batches that are seeded. We varied it at $ratio=50\%$ with all other hyperparameters fixed. Every interval from 2 to 16 outperforms on-policy training ($T=1$) by 2.1--2.9 points, while differing from one another by at most 0.7, with the best result at $T=8$. Across both ablations every seeded configuration, including the weakest (49.67 at $ratio=90\%$), scores above on-policy training and both off-policy baselines.

\section{Analysis}
\label{sec:analysis}

We now examine the trained runs themselves: where entropy and clipping concentrate once training restarts from seeds, which tokens the two truncation rules cut at, and what the extra generation costs.

\begin{figure}[t]
    \centering
    \includegraphics[width=0.76\columnwidth]{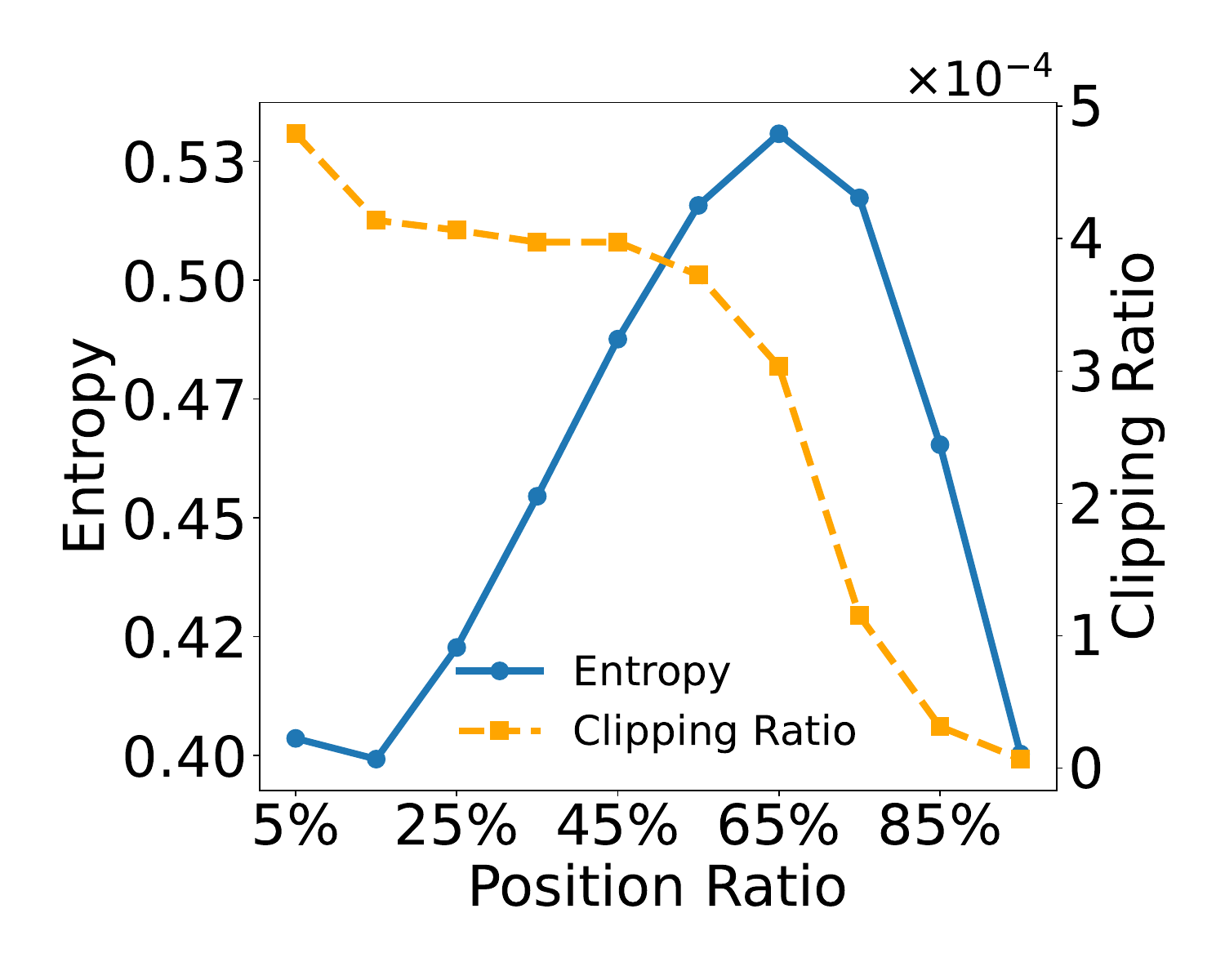}
    \caption{Entropy and clipping ratio vs.\ relative token position for the state${+}$action variant.}
    \label{fig:ts7}
\end{figure}

\subsection{Entropy at the Takeover Point}

Figure~\ref{fig:ts7} repeats the per-position analysis of Section~\ref{sec:entropy_collapse} for the state${+}$action variant ($ratio=70\%$, $T=16$); positions are normalized by the final sequence length, so the takeover point is spread around 70\%. Whereas entropy decreases along the sequence in the on-policy run (Figure~\ref{fig:on_wo_mini}), it peaks near the takeover point in the seeded run. This is consistent with the restart interpretation: after receiving a historical prefix, the current policy encounters a context in which several continuations remain plausible, that is, the change of continuation states described by Proposition~\ref{prop:coverage}. Clipping instead follows the seed--continuation split, concentrating on the seed tokens that are trained through their ratios to $\pi_b$ and vanishing at the end of the sequence; under mini-batch updates (Figure~\ref{fig:on_w_mini}) the two profiles coincided rather than separated. The state variant removes the seed-token contributions, so this picture describes the state${+}$action objective only. Appendix~\ref{sec:training_dynamics} shows the training reward curves.

\subsection{Truncation Point Analysis}
\label{sec:truncation_analysis}



\begin{figure}[t]
    \centering
    \begin{subfigure}[b]{0.78\columnwidth}
        \centering
        \includegraphics[width=\linewidth]{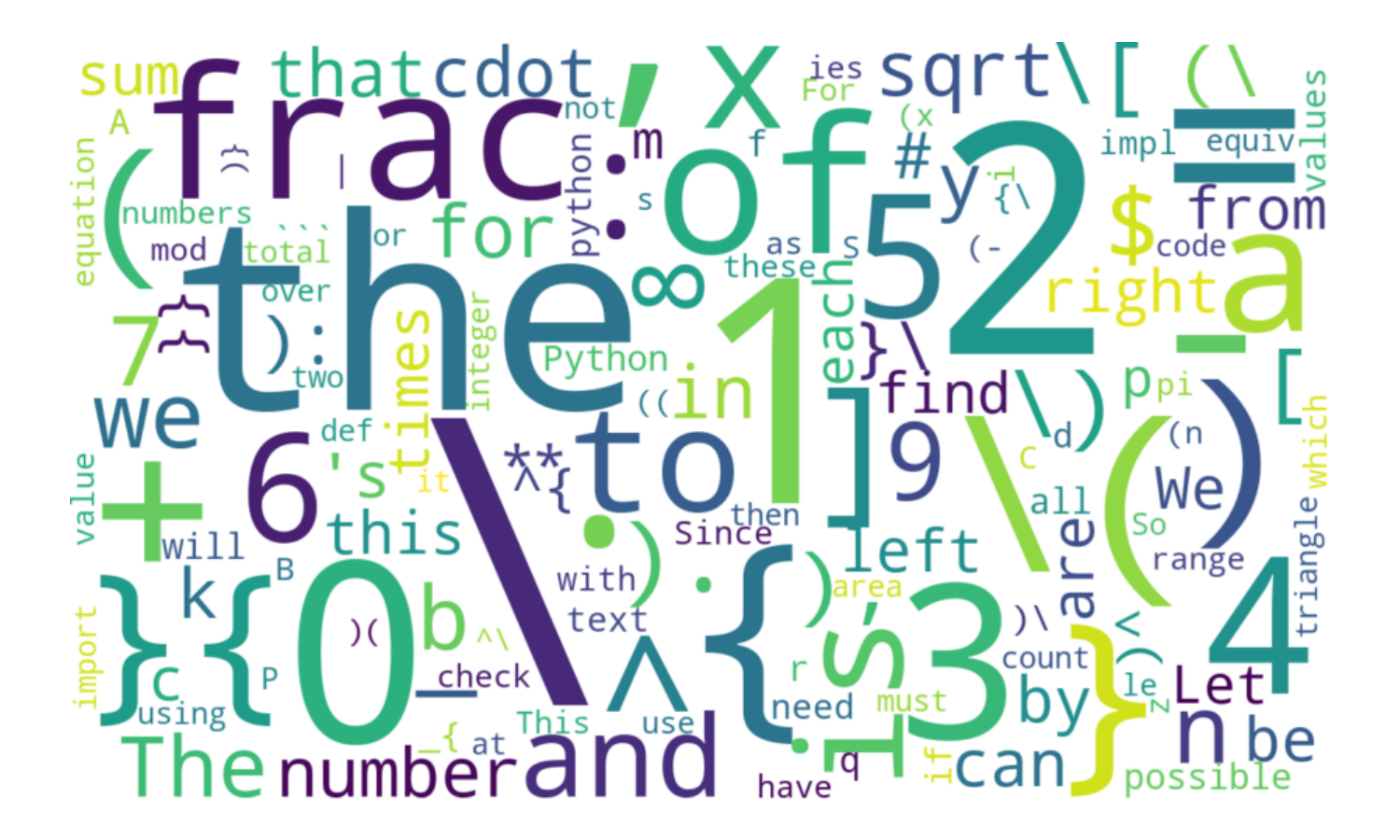}
        \caption{Length-ratio-based truncation.}
    \end{subfigure}

    \vspace{0.2em}
    \begin{subfigure}[b]{0.78\columnwidth}
        \centering
        \includegraphics[width=\linewidth]{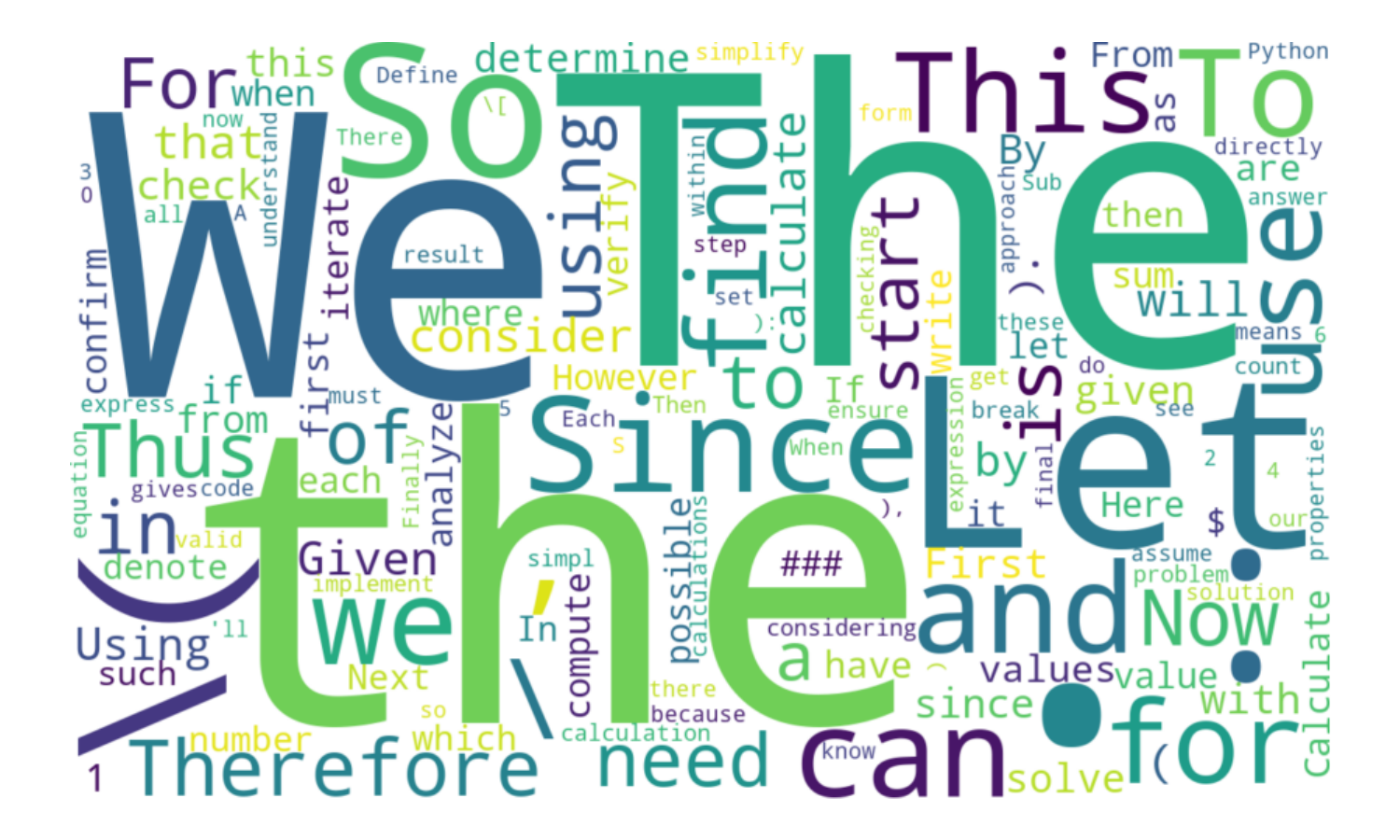}
        \caption{Entropy-based truncation.}
    \end{subfigure}
    \caption{Tokens at the truncation points under the two truncation strategies.}
    \label{fig:woldfreq}
\end{figure}

The truncation point is where the on-policy continuation takes over, so it determines the context from which the suffix is generated. Figure~\ref{fig:woldfreq} shows the tokens at these points. Entropy-based truncation concentrates on sentence-initial discourse words such as ``So,'' ``Since,'' ``Let,'' and ``Therefore,'' consistent with prior findings that high-entropy tokens in reasoning traces are often transitional words~\citep{DBLP:journals/corr/abs-2506-01939, DBLP:journals/corr/abs-2506-02867}. Length-ratio-based truncation is content-independent and cuts at more heterogeneous tokens, including numbers, operators, and LaTeX commands such as \texttt{\textbackslash frac}. Its continuations can therefore begin inside a computation, while entropy-based continuations more often begin at a step boundary. This qualitatively explains the result in Table~\ref{tab:main_results}: the length-ratio variant is stronger on both backbones, with the largest gap on DeepSeek-R1-Distill-Qwen-1.5B.

\subsection{Computational Overhead}

Thinking Seeds adds no resident model and no extra weight synchronization to the inference engine: the seed token ids and, for the state${+}$action variant, their behavior log-probabilities are cached when the rollouts are generated. The additional cost is generation. Because batches $2,\dots,T$ of each refresh cycle are completed by the current policy after truncation, the number of generated tokens grows by a factor of $1+(1-ratio)\frac{T-1}{T}$, i.e., $1.44\times$ at $ratio=50\%$ and $T=8$, while the measured wall-clock time per training step is $1.1$--$1.2\times$ that of on-policy training; Appendix~\ref{sec:overhead_details} explains the gap.

\section{Conclusion}

On-policy RL learns only from the reasoning states reached by the current policy. Thinking Seeds adds states from recent checkpoints by truncating their rollouts and letting the current policy continue from the prefix. At the population level the continuation has the policy-gradient form of a restart objective, while the prefix either remains in the loss through behavior-policy ratios or serves only as context.

The length-ratio variant improves over on-policy training and the two off-policy baselines on both mathematical backbones. On code generation, without retuning the mathematical configuration, LiveCodeBench-v6 rises from 48.58 to 50.20 for state${+}$action and 52.19 for state training. The position analysis shows an entropy peak near the takeover point and clipping concentrated on seed tokens. These results support historical prefixes as a practical way to change the start-state distribution.

\section*{Limitations}

Each configuration was trained once. We therefore do not report training-seed variance, and the smaller gaps, such as 54.45 versus 53.77 on DeepSeek-R1-Distill-Qwen-1.5B, may lie within run-to-run variation.

The mathematical experiments and both ablations use the state${+}$action variant. Its prefix-action term is subject to the context mismatch in Proposition~\ref{prop:bias}, and our analysis does not bound the dependence of its reward on the complete behavior-generated prefix. The state variant removes this direct term, but we evaluate it only on code generation. The coverage interpretation is qualitative: the relevant better-policy state distribution cannot be measured for an LLM, and Proposition~\ref{prop:coverage} describes a cycle with fixed parameters while the parameters change during training.

The seed distribution is fixed by design. Seeds come from the checkpoint at the start of each refresh cycle and are truncated by a fixed rule, which controls staleness but does not select the states most useful for restarting. Adaptive selection of seed sources or truncation points, for example using the current policy's success rate from a state, remains future work.


\bibliography{custom}

@inproceedings{DBLP:conf/nips/ChristianoLBMLA17,
  author       = {Paul F. Christiano and
                  Jan Leike and
                  Tom B. Brown and
                  Miljan Martic and
                  Shane Legg and
                  Dario Amodei},
  editor       = {Isabelle Guyon and
                  Ulrike von Luxburg and
                  Samy Bengio and
                  Hanna M. Wallach and
                  Rob Fergus and
                  S. V. N. Vishwanathan and
                  Roman Garnett},
  title        = {Deep Reinforcement Learning from Human Preferences},
  booktitle    = {Advances in Neural Information Processing Systems 30: Annual Conference
                  on Neural Information Processing Systems 2017, December 4-9, 2017,
                  Long Beach, CA, {USA}},
  pages        = {4299--4307},
  year         = {2017},
  url          = {https://proceedings.neurips.cc/paper/2017/hash/d5e2c0adad503c91f91df240d0cd4e49-Abstract.html},
  bibsource    = {dblp computer science bibliography, https://dblp.org}
}

@inproceedings{DBLP:conf/nips/StiennonO0ZLVRA20,
  author       = {Nisan Stiennon and
                  Long Ouyang and
                  Jeffrey Wu and
                  Daniel M. Ziegler and
                  Ryan Lowe and
                  Chelsea Voss and
                  Alec Radford and
                  Dario Amodei and
                  Paul F. Christiano},
  editor       = {Hugo Larochelle and
                  Marc'Aurelio Ranzato and
                  Raia Hadsell and
                  Maria{-}Florina Balcan and
                  Hsuan{-}Tien Lin},
  title        = {Learning to summarize with human feedback},
  booktitle    = {Advances in Neural Information Processing Systems 33: Annual Conference
                  on Neural Information Processing Systems 2020, NeurIPS 2020, December
                  6-12, 2020, virtual},
  year         = {2020},
  url          = {https://proceedings.neurips.cc/paper/2020/hash/1f89885d556929e98d3ef9b86448f951-Abstract.html},
  bibsource    = {dblp computer science bibliography, https://dblp.org}
}

@inproceedings{DBLP:conf/nips/Ouyang0JAWMZASR22,
  author       = {Long Ouyang and
                  Jeffrey Wu and
                  Xu Jiang and
                  Diogo Almeida and
                  Carroll L. Wainwright and
                  Pamela Mishkin and
                  Chong Zhang and
                  Sandhini Agarwal and
                  Katarina Slama and
                  Alex Ray and
                  John Schulman and
                  Jacob Hilton and
                  Fraser Kelton and
                  Luke Miller and
                  Maddie Simens and
                  Amanda Askell and
                  Peter Welinder and
                  Paul F. Christiano and
                  Jan Leike and
                  Ryan Lowe},
  editor       = {Sanmi Koyejo and
                  S. Mohamed and
                  A. Agarwal and
                  Danielle Belgrave and
                  K. Cho and
                  A. Oh},
  title        = {Training language models to follow instructions with human feedback},
  booktitle    = {Advances in Neural Information Processing Systems 35: Annual Conference
                  on Neural Information Processing Systems 2022, NeurIPS 2022, New Orleans,
                  LA, USA, November 28 - December 9, 2022},
  year         = {2022},
  url          = {http://papers.nips.cc/paper\_files/paper/2022/hash/b1efde53be364a73914f58805a001731-Abstract-Conference.html},
  bibsource    = {dblp computer science bibliography, https://dblp.org}
}

@article{DBLP:journals/corr/abs-2204-05862,
  author       = {Yuntao Bai and
                  Andy Jones and
                  Kamal Ndousse and
                  Amanda Askell and
                  Anna Chen and
                  Nova DasSarma and
                  Dawn Drain and
                  Stanislav Fort and
                  Deep Ganguli and
                  Tom Henighan and
                  Nicholas Joseph and
                  Saurav Kadavath and
                  Jackson Kernion and
                  Tom Conerly and
                  Sheer El Showk and
                  Nelson Elhage and
                  Zac Hatfield{-}Dodds and
                  Danny Hernandez and
                  Tristan Hume and
                  Scott Johnston and
                  Shauna Kravec and
                  Liane Lovitt and
                  Neel Nanda and
                  Catherine Olsson and
                  Dario Amodei and
                  Tom B. Brown and
                  Jack Clark and
                  Sam McCandlish and
                  Chris Olah and
                  Benjamin Mann and
                  Jared Kaplan},
  title        = {Training a Helpful and Harmless Assistant with Reinforcement Learning
                  from Human Feedback},
  journal      = {CoRR},
  volume       = {abs/2204.05862},
  year         = {2022},
  url          = {https://doi.org/10.48550/arXiv.2204.05862},
  doi          = {10.48550/ARXIV.2204.05862},
  eprinttype    = {arXiv},
  eprint       = {2204.05862},
  bibsource    = {dblp computer science bibliography, https://dblp.org}
}

@article{DBLP:journals/corr/abs-2411-15124,
  author       = {Nathan Lambert and
                  Jacob Morrison and
                  Valentina Pyatkin and
                  Shengyi Huang and
                  Hamish Ivison and
                  Faeze Brahman and
                  Lester James V. Miranda and
                  Alisa Liu and
                  Nouha Dziri and
                  Shane Lyu and
                  Yuling Gu and
                  Saumya Malik and
                  Victoria Graf and
                  Jena D. Hwang and
                  Jiangjiang Yang and
                  Ronan Le Bras and
                  Oyvind Tafjord and
                  Chris Wilhelm and
                  Luca Soldaini and
                  Noah A. Smith and
                  Yizhong Wang and
                  Pradeep Dasigi and
                  Hannaneh Hajishirzi},
  title        = {{T{\"{U}}LU} 3: Pushing Frontiers in Open Language Model Post-Training},
  journal      = {CoRR},
  volume       = {abs/2411.15124},
  year         = {2024},
  url          = {https://doi.org/10.48550/arXiv.2411.15124},
  doi          = {10.48550/ARXIV.2411.15124},
  eprinttype    = {arXiv},
  eprint       = {2411.15124},
  bibsource    = {dblp computer science bibliography, https://dblp.org}
}

@article{DBLP:journals/corr/abs-2402-03300,
  author       = {Zhihong Shao and
                  Peiyi Wang and
                  Qihao Zhu and
                  Runxin Xu and
                  Junxiao Song and
                  Mingchuan Zhang and
                  Y. K. Li and
                  Y. Wu and
                  Daya Guo},
  title        = {{DeepSeekMath:} Pushing the Limits of Mathematical Reasoning in Open
                  Language Models},
  journal      = {CoRR},
  volume       = {abs/2402.03300},
  year         = {2024},
  url          = {https://doi.org/10.48550/arXiv.2402.03300},
  doi          = {10.48550/ARXIV.2402.03300},
  eprinttype    = {arXiv},
  eprint       = {2402.03300},
  bibsource    = {dblp computer science bibliography, https://dblp.org}
}

@inproceedings{DBLP:conf/nips/HendrycksBKABTS21,
  author       = {Dan Hendrycks and
                  Collin Burns and
                  Saurav Kadavath and
                  Akul Arora and
                  Steven Basart and
                  Eric Tang and
                  Dawn Song and
                  Jacob Steinhardt},
  editor       = {Joaquin Vanschoren and
                  Sai{-}Kit Yeung},
  title        = {Measuring Mathematical Problem Solving With the {MATH} Dataset},
  booktitle    = {Proceedings of the Neural Information Processing Systems Track on
                  Datasets and Benchmarks 1, NeurIPS Datasets and Benchmarks 2021, December
                  2021, virtual},
  year         = {2021},
  url          = {https://datasets-benchmarks-proceedings.neurips.cc/paper/2021/hash/be83ab3ecd0db773eb2dc1b0a17836a1-Abstract-round2.html},
  bibsource    = {dblp computer science bibliography, https://dblp.org}
}

@inproceedings{DBLP:conf/icml/SchulmanLAJM15,
  author       = {John Schulman and
                  Sergey Levine and
                  Pieter Abbeel and
                  Michael I. Jordan and
                  Philipp Moritz},
  editor       = {Francis R. Bach and
                  David M. Blei},
  title        = {Trust Region Policy Optimization},
  booktitle    = {Proceedings of the 32nd International Conference on Machine Learning,
                  {ICML} 2015, Lille, France, 6-11 July 2015},
  series       = {{JMLR} Workshop and Conference Proceedings},
  volume       = {37},
  pages        = {1889--1897},
  publisher    = {JMLR.org},
  year         = {2015},
  url          = {http://proceedings.mlr.press/v37/schulman15.html},
  bibsource    = {dblp computer science bibliography, https://dblp.org}
}

@inproceedings{DBLP:conf/icml/LiXZL00L24,
  author       = {Ziniu Li and
                  Tian Xu and
                  Yushun Zhang and
                  Zhihang Lin and
                  Yang Yu and
                  Ruoyu Sun and
                  Zhi{-}Quan Luo},
  title        = {{ReMax:} {A} Simple, Effective, and Efficient Reinforcement Learning
                  Method for Aligning Large Language Models},
  booktitle    = {Forty-first International Conference on Machine Learning, {ICML} 2024,
                  Vienna, Austria, July 21-27, 2024},
  publisher    = {OpenReview.net},
  year         = {2024},
  url          = {https://openreview.net/forum?id=Stn8hXkpe6},
  bibsource    = {dblp computer science bibliography, https://dblp.org}
}

@inproceedings{DBLP:conf/acl/AhmadianCGFKPUH24,
  author       = {Arash Ahmadian and
                  Chris Cremer and
                  Matthias Gall{\'{e}} and
                  Marzieh Fadaee and
                  Julia Kreutzer and
                  Olivier Pietquin and
                  Ahmet {\"{U}}st{\"{u}}n and
                  Sara Hooker},
  editor       = {Lun{-}Wei Ku and
                  Andre Martins and
                  Vivek Srikumar},
  title        = {Back to Basics: Revisiting {REINFORCE}-Style Optimization for Learning
                  from Human Feedback in {LLMs}},
  booktitle    = {Proceedings of the 62nd Annual Meeting of the Association for Computational
                  Linguistics (Volume 1: Long Papers), {ACL} 2024, Bangkok, Thailand,
                  August 11-16, 2024},
  pages        = {12248--12267},
  publisher    = {Association for Computational Linguistics},
  year         = {2024},
  url          = {https://doi.org/10.18653/v1/2024.acl-long.662},
  doi          = {10.18653/V1/2024.ACL-LONG.662},
  bibsource    = {dblp computer science bibliography, https://dblp.org}
}

@article{DBLP:journals/corr/abs-2501-03262,
  author       = {Jian Hu and
                  Jason Klein Liu and
                  Haotian Xu and
                  Wei Shen},
  title        = {{REINFORCE++:} Stabilizing Critic-Free Policy Optimization with
                  Global Advantage Normalization},
  journal      = {CoRR},
  volume       = {abs/2501.03262},
  year         = {2025},
  url          = {https://doi.org/10.48550/arXiv.2501.03262},
  doi          = {10.48550/ARXIV.2501.03262},
  eprinttype    = {arXiv},
  eprint       = {2501.03262},
  bibsource    = {dblp computer science bibliography, https://dblp.org}
}

@article{DBLP:journals/corr/abs-2504-07912,
  author       = {Rosie Zhao and
                  Alexandru Meterez and
                  Sham M. Kakade and
                  Cengiz Pehlevan and
                  Samy Jelassi and
                  Eran Malach},
  title        = {Echo Chamber: {RL} Post-training Amplifies Behaviors Learned in Pretraining},
  journal      = {CoRR},
  volume       = {abs/2504.07912},
  year         = {2025},
  url          = {https://doi.org/10.48550/arXiv.2504.07912},
  doi          = {10.48550/ARXIV.2504.07912},
  eprinttype    = {arXiv},
  eprint       = {2504.07912},
  bibsource    = {dblp computer science bibliography, https://dblp.org}
}

@inproceedings{DBLP:journals/corr/abs-2504-13837,
  author       = {Yue, Yang and Chen, Zhiqi and Lu, Rui and Zhao, Andrew and Wang, Zhaokai and
                  Yue, Yang and Song, Shiji and Huang, Gao},
  title        = {Does Reinforcement Learning Really Incentivize Reasoning Capacity
                  in {LLMs} Beyond the Base Model?},
  editor       = {D. Belgrave and C. Zhang and H. Lin and R. Pascanu and
                  P. Koniusz and M. Ghassemi and N. Chen},
  booktitle    = {Advances in Neural Information Processing Systems},
  publisher    = {Curran Associates, Inc.},
  volume       = {38, Main Conference},
  year         = {2025},
  pages        = {57654--57689},
  url          = {https://papers.nips.cc/paper_files/paper/2025/file/537d5aa768c2d534016a4d06f87bc8fb-Paper-Conference.pdf},
  doi          = {10.52202/085713-1933}
}

@article{DBLP:journals/corr/abs-2503-01307,
  author       = {Kanishk Gandhi and
                  Ayush Chakravarthy and
                  Anikait Singh and
                  Nathan Lile and
                  Noah D. Goodman},
  title        = {Cognitive Behaviors that Enable Self-Improving Reasoners, or, Four
                  Habits of Highly Effective {STaRs}},
  journal      = {CoRR},
  volume       = {abs/2503.01307},
  year         = {2025},
  url          = {https://doi.org/10.48550/arXiv.2503.01307},
  doi          = {10.48550/ARXIV.2503.01307},
  eprinttype    = {arXiv},
  eprint       = {2503.01307},
  bibsource    = {dblp computer science bibliography, https://dblp.org}
}

@inproceedings{DBLP:journals/corr/abs-2504-14945,
  author       = {Yan, Jianhao and Li, Yafu and Hu, Zican and Wang, Zhi and Cui, Ganqu and
                  Qu, Xiaoye and Cheng, Yu and Zhang, Yue},
  title        = {Learning to Reason under Off-Policy Guidance},
  editor       = {D. Belgrave and C. Zhang and H. Lin and R. Pascanu and
                  P. Koniusz and M. Ghassemi and N. Chen},
  booktitle    = {Advances in Neural Information Processing Systems},
  publisher    = {Curran Associates, Inc.},
  volume       = {38, Main Conference},
  year         = {2025},
  pages        = {129799--129828},
  url          = {https://doi.org/10.52202/085713-3904},
  doi          = {10.52202/085713-3904}
}

@article{DBLP:journals/corr/abs-2510-01161,
  author       = {Haizhong Zheng and
                  Jiawei Zhao and
                  Beidi Chen},
  title        = {Prosperity before Collapse: How Far Can Off-Policy {RL} Reach with
                  Stale Data on {LLMs}?},
  journal      = {CoRR},
  volume       = {abs/2510.01161},
  year         = {2025},
  url          = {https://doi.org/10.48550/arXiv.2510.01161},
  doi          = {10.48550/ARXIV.2510.01161},
  eprinttype    = {arXiv},
  eprint       = {2510.01161},
  bibsource    = {dblp computer science bibliography, https://dblp.org}
}

@article{DBLP:journals/corr/abs-2508-11016,
  author       = {Qingbin Li and
                  Rongkun Xue and
                  Jie Wang and
                  Ming Zhou and
                  Zhi Li and
                  Xiaofeng Ji and
                  Yongqi Wang and
                  Miao Liu and
                  Zheming Yang and
                  Minghui Qiu and
                  Jing Yang},
  title        = {{CURE:} Critical-Token-Guided Re-Concatenation for Entropy-Collapse
                  Prevention},
  journal      = {CoRR},
  volume       = {abs/2508.11016},
  year         = {2025},
  url          = {https://doi.org/10.48550/arXiv.2508.11016},
  doi          = {10.48550/ARXIV.2508.11016},
  eprinttype    = {arXiv},
  eprint       = {2508.11016},
  bibsource    = {dblp computer science bibliography, https://dblp.org}
}

@article{DBLP:journals/corr/abs-2505-22257,
  author       = {Youssef Mroueh and
                  Nicolas Dupuis and
                  Brian Belgodere and
                  Apoorva Nitsure and
                  Mattia Rigotti and
                  Kristjan H. Greenewald and
                  Jir{\'{\i}} Navr{\'{a}}til and
                  Jerret Ross and
                  Jesus Rios},
  title        = {Revisiting Group Relative Policy Optimization: Insights into On-Policy
                  and Off-Policy Training},
  journal      = {CoRR},
  volume       = {abs/2505.22257},
  year         = {2025},
  url          = {https://doi.org/10.48550/arXiv.2505.22257},
  doi          = {10.48550/ARXIV.2505.22257},
  eprinttype    = {arXiv},
  eprint       = {2505.22257},
  bibsource    = {dblp computer science bibliography, https://dblp.org}
}

@article{DBLP:journals/corr/abs-2506-21495,
  author       = {Jack Lanchantin and
                  Angelica Chen and
                  Janice Lan and
                  Xian Li and
                  Swarnadeep Saha and
                  Tianlu Wang and
                  Jing Xu and
                  Ping Yu and
                  Weizhe Yuan and
                  Jason E. Weston and
                  Sainbayar Sukhbaatar and
                  Ilia Kulikov},
  title        = {Bridging Offline and Online Reinforcement Learning for {LLMs}},
  journal      = {CoRR},
  volume       = {abs/2506.21495},
  year         = {2025},
  url          = {https://doi.org/10.48550/arXiv.2506.21495},
  doi          = {10.48550/ARXIV.2506.21495},
  eprinttype    = {arXiv},
  eprint       = {2506.21495},
  bibsource    = {dblp computer science bibliography, https://dblp.org}
}

@inproceedings{DBLP:journals/corr/abs-2503-14476,
  author       = {Yu, Qiying and Zhang, Zheng and Zhu, Ruofei and Yuan, Yufeng and Zuo, Xiaochen and
                  Yue, Yu and Dai, Weinan and Fan, Tiantian and Liu, Gaohong and
                  Liu, Juncai and Liu, LingJun and Liu, Xin and Lin, Haibin and
                  Lin, Zhiqi and Ma, Bole and Sheng, Guangming and Tong, Yuxuan and
                  Zhang, Chi and Zhang, Mofan and Zhang, Ru and Zhang, Wang and
                  Zhu, Hang and Zhu, Jinhua and Chen, Jiaze and Chen, Jiangjie and
                  Wang, Chengyi and Yu, Hongli and Song, Yuxuan and Wei, Xiangpeng and
                  Zhou, Hao and Liu, Jingjing and Ma, Wei-Ying and Zhang, Ya-Qin and
                  Yan, Lin and Wu, Yonghui and Wang, Mingxuan},
  title        = {{DAPO:} An Open-Source {LLM} Reinforcement Learning System at Scale},
  editor       = {D. Belgrave and C. Zhang and H. Lin and R. Pascanu and
                  P. Koniusz and M. Ghassemi and N. Chen},
  booktitle    = {Advances in Neural Information Processing Systems},
  publisher    = {Curran Associates, Inc.},
  volume       = {38, Main Conference},
  year         = {2025},
  pages        = {125532--125554},
  url          = {https://doi.org/10.52202/085713-3775},
  doi          = {10.52202/085713-3775}
}

@article{DBLP:journals/corr/abs-2409-12122,
  author       = {An Yang and
                  Beichen Zhang and
                  Binyuan Hui and
                  Bofei Gao and
                  Bowen Yu and
                  Chengpeng Li and
                  Dayiheng Liu and
                  Jianhong Tu and
                  Jingren Zhou and
                  Junyang Lin and
                  Keming Lu and
                  Mingfeng Xue and
                  Runji Lin and
                  Tianyu Liu and
                  Xingzhang Ren and
                  Zhenru Zhang},
  title        = {{Qwen2.5-Math} Technical Report: Toward Mathematical Expert Model via
                  Self-Improvement},
  journal      = {CoRR},
  volume       = {abs/2409.12122},
  year         = {2024},
  url          = {https://doi.org/10.48550/arXiv.2409.12122},
  doi          = {10.48550/ARXIV.2409.12122},
  eprinttype    = {arXiv},
  eprint       = {2409.12122},
  bibsource    = {dblp computer science bibliography, https://dblp.org}
}

@article{DBLP:journals/corr/abs-2501-12948,
  author       = {DeepSeek{-}AI},
  title        = {{DeepSeek-R1} incentivizes reasoning in {LLMs} through reinforcement
                  learning},
  journal      = {Nature},
  volume       = {645},
  number       = {8081},
  pages        = {633--638},
  month        = sep,
  year         = {2025},
  url          = {https://doi.org/10.1038/s41586-025-09422-z},
  doi          = {10.1038/s41586-025-09422-z},
  issn         = {1476-4687}
}

@misc{aops_aime,
  author       = {{Art of Problem Solving}},
  title        = {{AIME} Problems and Solutions},
  howpublished = {\url{https://artofproblemsolving.com/wiki/index.php/AIME_Problems_and_Solutions}},
  year         = {2024},
  note         = {Accessed: 2026-10-13}
}

@misc{aops_amc,
  author       = {{Art of Problem Solving}},
  title        = {{AMC} Problems and Solutions},
  howpublished = {\url{https://artofproblemsolving.com/wiki/index.php?title=AMC_Problems_and_Solutions}},
  year         = {2024},
  note         = {Accessed: 2026-10-13}
}

@inproceedings{DBLP:conf/nips/LewkowyczADDMRS22,
  author       = {Aitor Lewkowycz and
                  Anders Andreassen and
                  David Dohan and
                  Ethan Dyer and
                  Henryk Michalewski and
                  Vinay V. Ramasesh and
                  Ambrose Slone and
                  Cem Anil and
                  Imanol Schlag and
                  Theo Gutman{-}Solo and
                  Yuhuai Wu and
                  Behnam Neyshabur and
                  Guy Gur{-}Ari and
                  Vedant Misra},
  editor       = {Sanmi Koyejo and
                  S. Mohamed and
                  A. Agarwal and
                  Danielle Belgrave and
                  K. Cho and
                  A. Oh},
  title        = {Solving Quantitative Reasoning Problems with Language Models},
  booktitle    = {Advances in Neural Information Processing Systems 35: Annual Conference
                  on Neural Information Processing Systems 2022, NeurIPS 2022, New Orleans,
                  LA, USA, November 28 - December 9, 2022},
  year         = {2022},
  url          = {http://papers.nips.cc/paper\_files/paper/2022/hash/18abbeef8cfe9203fdf9053c9c4fe191-Abstract-Conference.html},
  bibsource    = {dblp computer science bibliography, https://dblp.org}
}

@inproceedings{DBLP:conf/acl/HeLBHTSHHHZLQL024,
  author       = {Chaoqun He and
                  Renjie Luo and
                  Yuzhuo Bai and
                  Shengding Hu and
                  Zhen Leng Thai and
                  Junhao Shen and
                  Jinyi Hu and
                  Xu Han and
                  Yujie Huang and
                  Yuxiang Zhang and
                  Jie Liu and
                  Lei Qi and
                  Zhiyuan Liu and
                  Maosong Sun},
  editor       = {Lun{-}Wei Ku and
                  Andre Martins and
                  Vivek Srikumar},
  title        = {{OlympiadBench:} {A} Challenging Benchmark for Promoting {AGI} with
                  Olympiad-Level Bilingual Multimodal Scientific Problems},
  booktitle    = {Proceedings of the 62nd Annual Meeting of the Association for Computational
                  Linguistics (Volume 1: Long Papers), {ACL} 2024, Bangkok, Thailand,
                  August 11-16, 2024},
  pages        = {3828--3850},
  publisher    = {Association for Computational Linguistics},
  year         = {2024},
  url          = {https://doi.org/10.18653/v1/2024.acl-long.211},
  doi          = {10.18653/V1/2024.ACL-LONG.211},
  bibsource    = {dblp computer science bibliography, https://dblp.org}
}

@article{DBLP:journals/corr/abs-2505-22617,
  author       = {Ganqu Cui and
                  Yuchen Zhang and
                  Jiacheng Chen and
                  Lifan Yuan and
                  Zhi Wang and
                  Yuxin Zuo and
                  Haozhan Li and
                  Yuchen Fan and
                  Huayu Chen and
                  Weize Chen and
                  Zhiyuan Liu and
                  Hao Peng and
                  Lei Bai and
                  Wanli Ouyang and
                  Yu Cheng and
                  Bowen Zhou and
                  Ning Ding},
  title        = {The Entropy Mechanism of Reinforcement Learning for Reasoning Language
                  Models},
  journal      = {CoRR},
  volume       = {abs/2505.22617},
  year         = {2025},
  url          = {https://doi.org/10.48550/arXiv.2505.22617},
  doi          = {10.48550/ARXIV.2505.22617},
  eprinttype    = {arXiv},
  eprint       = {2505.22617},
  bibsource    = {dblp computer science bibliography, https://dblp.org}
}

@article{DBLP:journals/corr/abs-2505-22312,
  author       = {Jujie He and
                  Jiacai Liu and
                  Chris Yuhao Liu and
                  Rui Yan and
                  Chaojie Wang and
                  Peng Cheng and
                  Xiaoyu Zhang and
                  Fuxiang Zhang and
                  Jiacheng Xu and
                  Wei Shen and
                  Siyuan Li and
                  Liang Zeng and
                  Tianwen Wei and
                  Cheng Cheng and
                  Bo An and
                  Yang Liu and
                  Yahui Zhou},
  title        = {Skywork Open Reasoner 1 Technical Report},
  journal      = {CoRR},
  volume       = {abs/2505.22312},
  year         = {2025},
  url          = {https://doi.org/10.48550/arXiv.2505.22312},
  doi          = {10.48550/ARXIV.2505.22312},
  eprinttype    = {arXiv},
  eprint       = {2505.22312},
  bibsource    = {dblp computer science bibliography, https://dblp.org}
}

@inproceedings{DBLP:journals/corr/abs-2505-24298,
  author       = {Fu, Wei and Gao, Jiaxuan and Shen, Xujie and Zhu, Chen and Mei, Zhiyu and
                  He, Chuyi and Xu, Shusheng and Wei, Guo and Mei, Jun and
                  Wang, Jiashu and Yang, Tongkai and Yuan, Binhang and Wu, Yi},
  title        = {{AReaL:} {A} Large-Scale Asynchronous Reinforcement Learning System
                  for Language Reasoning},
  editor       = {D. Belgrave and C. Zhang and H. Lin and R. Pascanu and
                  P. Koniusz and M. Ghassemi and N. Chen},
  booktitle    = {Advances in Neural Information Processing Systems},
  publisher    = {Curran Associates, Inc.},
  volume       = {38, Main Conference},
  year         = {2025},
  pages        = {36256--36282},
  url          = {https://papers.nips.cc/paper_files/paper/2025/file/33c00862bfa29ac72ecf630a41e19352-Paper-Conference.pdf},
  doi          = {10.52202/085713-1218}
}

@misc{slime_github,
  author       = {Zilin Zhu and Chengxing Xie and Xin Lv and slime Contributors},
  title        = {slime: An {LLM} post-training framework for {RL} scaling},
  year         = {2025},
  howpublished = {\url{https://github.com/THUDM/slime}},
  note         = {GitHub repository. Corresponding author: Xin Lv},
  urldate      = {2025-06-19}
}

@inproceedings{DBLP:conf/iclr/NoukhovitchHXHA25,
  author       = {Michael Noukhovitch and
                  Shengyi Huang and
                  Sophie Xhonneux and
                  Arian Hosseini and
                  Rishabh Agarwal and
                  Aaron C. Courville},
  title        = {Asynchronous {RLHF:} Faster and More Efficient Off-Policy {RL} for
                  Language Models},
  booktitle    = {The Thirteenth International Conference on Learning Representations,
                  {ICLR} 2025, Singapore, April 24-28, 2025},
  publisher    = {OpenReview.net},
  year         = {2025},
  url          = {https://openreview.net/forum?id=FhTAG591Ve},
  bibsource    = {dblp computer science bibliography, https://dblp.org}
}

@article{DBLP:journals/corr/abs-2504-15930,
  author       = {Yinmin Zhong and
                  Zili Zhang and
                  Xiaoniu Song and
                  Hanpeng Hu and
                  Chao Jin and
                  Bingyang Wu and
                  Nuo Chen and
                  Yukun Chen and
                  Yu Zhou and
                  Changyi Wan and
                  Hongyu Zhou and
                  Yimin Jiang and
                  Yibo Zhu and
                  Daxin Jiang},
  title        = {{StreamRL:} Scalable, Heterogeneous, and Elastic {RL} for {LLMs} with
                  Disaggregated Stream Generation},
  journal      = {CoRR},
  volume       = {abs/2504.15930},
  year         = {2025},
  url          = {https://doi.org/10.48550/arXiv.2504.15930},
  doi          = {10.48550/ARXIV.2504.15930},
  eprinttype    = {arXiv},
  eprint       = {2504.15930},
  bibsource    = {dblp computer science bibliography, https://dblp.org}
}

@article{DBLP:journals/corr/abs-2508-18588,
  author       = {Jingkai He and
                  Tianjian Li and
                  Erhu Feng and
                  Dong Du and
                  Qian Liu and
                  Tao Liu and
                  Yubin Xia and
                  Haibo Chen},
  title        = {History Rhymes: Accelerating {LLM} Reinforcement Learning with {RhymeRL}},
  journal      = {CoRR},
  volume       = {abs/2508.18588},
  year         = {2025},
  url          = {https://doi.org/10.48550/arXiv.2508.18588},
  doi          = {10.48550/ARXIV.2508.18588},
  eprinttype    = {arXiv},
  eprint       = {2508.18588},
  bibsource    = {dblp computer science bibliography, https://dblp.org}
}

@inproceedings{DBLP:journals/corr/abs-2507-02841,
    title = "{S}tep{H}int: Multi-level Stepwise Hints Enhance Reinforcement Learning to Reason",
    author = "Zhang, Kaiyi  and
      Lv, Ang  and
      Li, Jinpeng  and
      Wang, Yongbo  and
      Wang, Feng  and
      Hu, Haoyuan  and
      Yan, Rui",
    editor = "Liakata, Maria  and
      Moreira, Viviane P.  and
      Zhang, Jiajun  and
      Jurgens, David",
    booktitle = "Proceedings of the 64th Annual Meeting of the {A}ssociation for {C}omputational {L}inguistics (Volume 1: Long Papers)",
    month = jul,
    year = "2026",
    address = "San Diego, California, United States",
    publisher = "Association for Computational Linguistics",
    url = "https://aclanthology.org/2026.acl-long.1755/",
    doi = "10.18653/v1/2026.acl-long.1755",
    pages = "37846--37864",
    ISBN = "979-8-89176-390-6"
}

@inproceedings{DBLP:journals/corr/abs-2506-01939,
  author       = {Wang, Shenzhi and Yu, Le and Gao, Chang and Zheng, Chujie and Liu, Shixuan and
                  Lu, Rui and Dang, Kai and Chen, Xiong-Hui and Yang, Jianxin and
                  Zhang, Zhenru and Liu, Yuqiong and Yang, An and Zhao, Andrew and
                  Yue, Yang and Song, Shiji and Yu, Bowen and Huang, Gao and
                  Lin, Junyang},
  title        = {Beyond the 80/20 Rule: {High-Entropy} Minority Tokens Drive Effective
                  Reinforcement Learning for {LLM} Reasoning},
  editor       = {D. Belgrave and C. Zhang and H. Lin and R. Pascanu and
                  P. Koniusz and M. Ghassemi and N. Chen},
  booktitle    = {Advances in Neural Information Processing Systems},
  publisher    = {Curran Associates, Inc.},
  volume       = {38, Main Conference},
  year         = {2025},
  pages        = {127985--128019},
  url          = {https://doi.org/10.52202/085713-3850},
  doi          = {10.52202/085713-3850}
}

@inproceedings{DBLP:journals/corr/abs-2506-02867,
  author       = {Qian, Chen and Liu, Dongrui and Wen, Haochen and Bai, Zhen and
                  Liu, Yong and Shao, Jing},
  title        = {Demystifying Reasoning Dynamics with Mutual Information: Thinking
                  Tokens are Information Peaks in {LLM} Reasoning},
  editor       = {D. Belgrave and C. Zhang and H. Lin and R. Pascanu and
                  P. Koniusz and M. Ghassemi and N. Chen},
  booktitle    = {Advances in Neural Information Processing Systems},
  publisher    = {Curran Associates, Inc.},
  volume       = {38, Main Conference},
  year         = {2025},
  pages        = {12533--12572},
  url          = {https://papers.nips.cc/paper_files/paper/2025/file/12867c8f1888d97d4e9f95f03c0f3ec9-Paper-Conference.pdf},
  doi          = {10.52202/085713-0421}
}

@article{DBLP:journals/corr/abs-2505-23585,
  author       = {Yaru Hao and
                  Li Dong and
                  Xun Wu and
                  Shaohan Huang and
                  Zewen Chi and
                  Furu Wei},
  title        = {On-Policy {RL} with Optimal Reward Baseline},
  journal      = {CoRR},
  volume       = {abs/2505.23585},
  year         = {2025},
  url          = {https://doi.org/10.48550/arXiv.2505.23585},
  doi          = {10.48550/ARXIV.2505.23585},
  eprinttype    = {arXiv},
  eprint       = {2505.23585},
  bibsource    = {dblp computer science bibliography, https://dblp.org}
}

@article{DBLP:journals/corr/abs-2509-23232,
  author       = {Bingshuai Liu and
                  Ante Wang and
                  Zijun Min and
                  Liang Yao and
                  Haibo Zhang and
                  Yang Liu and
                  Anxiang Zeng and
                  Jinsong Su},
  title        = {{SPEC-RL:} Accelerating On-Policy Reinforcement Learning with Speculative
                  Rollouts},
  journal      = {CoRR},
  volume       = {abs/2509.23232},
  year         = {2025},
  url          = {https://doi.org/10.48550/arXiv.2509.23232},
  doi          = {10.48550/ARXIV.2509.23232},
  eprinttype    = {arXiv},
  eprint       = {2509.23232},
  bibsource    = {dblp computer science bibliography, https://dblp.org}
}

@article{DBLP:journals/corr/abs-2506-05433,
  author       = {Zikang Liu and
                  Tongtian Yue and
                  Yepeng Tang and
                  Longteng Guo and
                  Junxian Cai and
                  Qingbin Liu and
                  Xi Chen and
                  Jing Liu},
  title        = {Prefix Grouper: Efficient {GRPO} Training through Shared-Prefix Forward},
  journal      = {CoRR},
  volume       = {abs/2506.05433},
  year         = {2025},
  url          = {https://doi.org/10.48550/arXiv.2506.05433},
  doi          = {10.48550/ARXIV.2506.05433},
  eprinttype    = {arXiv},
  eprint       = {2506.05433},
  bibsource    = {dblp computer science bibliography, https://dblp.org}
}

@article{DBLP:journals/corr/abs-2503-20783,
  author       = {Zichen Liu and
                  Changyu Chen and
                  Wenjun Li and
                  Penghui Qi and
                  Tianyu Pang and
                  Chao Du and
                  Wee Sun Lee and
                  Min Lin},
  title        = {Understanding {R1}-Zero-Like Training: {A} Critical Perspective},
  journal      = {CoRR},
  volume       = {abs/2503.20783},
  year         = {2025},
  url          = {https://doi.org/10.48550/arXiv.2503.20783},
  doi          = {10.48550/ARXIV.2503.20783},
  eprinttype    = {arXiv},
  eprint       = {2503.20783},
  bibsource    = {dblp computer science bibliography, https://dblp.org}
}

@article{DBLP:journals/corr/abs-2405-08448,
  author       = {Yunhao Tang and
                  Zhaohan Daniel Guo and
                  Zeyu Zheng and
                  Daniele Calandriello and
                  Yuan Cao and
                  Eugene Tarassov and
                  R{\'{e}}mi Munos and
                  Bernardo {\'{A}}vila Pires and
                  Michal Valko and
                  Yong Cheng and
                  Will Dabney},
  title        = {Understanding the performance gap between online and offline alignment
                  algorithms},
  journal      = {CoRR},
  volume       = {abs/2405.08448},
  year         = {2024},
  url          = {https://doi.org/10.48550/arXiv.2405.08448},
  doi          = {10.48550/ARXIV.2405.08448},
  eprinttype    = {arXiv},
  eprint       = {2405.08448},
  bibsource    = {dblp computer science bibliography, https://dblp.org}
}

@article{DBLP:journals/corr/abs-2509-06923,
  author       = {Ziheng Li and
                  Zexu Sun and
                  Jinman Zhao and
                  Erxue Min and
                  Yongcheng Zeng and
                  Hui Wu and
                  Hengyi Cai and
                  Shuaiqiang Wang and
                  Dawei Yin and
                  Xu Chen and
                  Zhi{-}Hong Deng},
  title        = {Staying in the Sweet Spot: Responsive Reasoning Evolution via Capability-Adaptive
                  Hint Scaffolding},
  journal      = {CoRR},
  volume       = {abs/2509.06923},
  year         = {2025},
  url          = {https://doi.org/10.48550/arXiv.2509.06923},
  doi          = {10.48550/ARXIV.2509.06923},
  eprinttype    = {arXiv},
  eprint       = {2509.06923},
  bibsource    = {dblp computer science bibliography, https://dblp.org}
}

@article{zhang2025adhint,
  title={{ADHint}: Adaptive Hints with Difficulty Priors for Reinforcement Learning},
  author={Zhang, Feng and Tan, Zezhong and Ma, Xinhong and Dong, Ziqiang and Leng, Xi and Zhao, Jianfei and Sun, Xin and Yang, Yang and Jiang, Guanjun},
  journal={arXiv preprint arXiv:2512.13095},
  year={2025}
}

@article{huang2025blending,
  title={Blending supervised and reinforcement fine-tuning with prefix sampling},
  author={Huang, Zeyu and Cheng, Tianhao and Qiu, Zihan and Wang, Zili and Xu, Yinghui and Ponti, Edoardo M and Titov, Ivan},
  journal={arXiv preprint arXiv:2507.01679},
  year={2025}
}

@inproceedings{chen2025lppo,
    title = "From Data-Centric to Sample-Centric: Enhancing {LLM} Reasoning via Progressive Optimization",
    author = "Chen, Xinjie  and
      Liao, Minpeng  and
      Chen, Guoxin  and
      Li, Chengxi  and
      Fu, Biao  and
      Fan, Kai  and
      Liu, Xinggao",
    editor = "Liakata, Maria  and
      Moreira, Viviane P.  and
      Zhang, Jiajun  and
      Jurgens, David",
    booktitle = "Proceedings of the 64th Annual Meeting of the {A}ssociation for {C}omputational {L}inguistics (Volume 1: Long Papers)",
    month = jul,
    year = "2026",
    address = "San Diego, California, United States",
    publisher = "Association for Computational Linguistics",
    url = "https://aclanthology.org/2026.acl-long.101/",
    doi = "10.18653/v1/2026.acl-long.101",
    pages = "2227--2242",
    ISBN = "979-8-89176-390-6"
}

@article{wu2025dft,
  title={On the generalization of {SFT}: {A} reinforcement learning perspective with reward rectification},
  author={Wu, Yongliang and Zhou, Yizhou and Zhou, Ziheng and Peng, Yingzhe and Ye, Xinyu and Hu, Xinting and Zhu, Wenbo and Qi, Lu and Yang, Ming-Hsuan and Yang, Xu},
  journal={arXiv preprint arXiv:2508.05629},
  year={2025}
}

@article{zhang2025chord,
  title={On-Policy {RL} Meets Off-Policy Experts: Harmonizing Supervised Fine-Tuning and Reinforcement Learning via Dynamic Weighting},
  author={Zhang, Wenhao and Xie, Yuexiang and Sun, Yuchang and Chen, Yanxi and Wang, Guoyin and Li, Yaliang and Ding, Bolin and Zhou, Jingren},
  journal={arXiv preprint arXiv:2508.11408},
  year={2025}
}

@article{yang2025qwen3,
  title={Qwen3 Technical Report},
  author={Yang, An and Li, Anfeng and Yang, Baosong and Zhang, Beichen and Hui, Binyuan and others},
  journal={arXiv preprint arXiv:2505.09388},
  year={2025}
}

@inproceedings{jain2025livecodebench,
  title={{LiveCodeBench}: Holistic and Contamination Free Evaluation of Large Language Models for Code},
  author={Jain, Naman and Han, King and Gu, Alex and Li, Wen-Ding and Yan, Fanjia and Zhang, Tianjun and Wang, Sida and Solar-Lezama, Armando and Sen, Koushik and Stoica, Ion},
  booktitle={The Thirteenth International Conference on Learning Representations},
  year={2025}
}

@article{su2025klear,
  title={{Klear-Reasoner}: Advancing Reasoning Capability via Gradient-Preserving Clipping Policy Optimization},
  author={Su, Zhenpeng and Pan, Leiyu and Bai, Xue and Liu, Dening and Dong, Guanting and Huang, Jiaming and Lv, Minxuan and Hu, Wenping and Zhang, Fuzheng and Gai, Kun and Zhou, Guorui},
  journal={arXiv preprint arXiv:2508.07629},
  year={2025}
}

@inproceedings{kakade2002approximately,
  title={Approximately Optimal Approximate Reinforcement Learning},
  author={Kakade, Sham and Langford, John},
  booktitle={Proceedings of the Nineteenth International Conference on Machine Learning (ICML)},
  pages={267--274},
  year={2002}
}

@article{agarwal2021theory,
  title={On the Theory of Policy Gradient Methods: Optimality, Approximation, and Distribution Shift},
  author={Agarwal, Alekh and Kakade, Sham M. and Lee, Jason D. and Mahajan, Gaurav},
  journal={Journal of Machine Learning Research},
  volume={22},
  number={98},
  pages={1--76},
  year={2021}
}

@inproceedings{liu2018breaking,
  title     = {Breaking the Curse of Horizon: Infinite-Horizon Off-Policy Estimation},
  author    = {Liu, Qiang and Li, Lihong and Tang, Ziyang and Zhou, Dengyong},
  booktitle = {Advances in Neural Information Processing Systems},
  volume    = {31},
  pages     = {5356--5366},
  year      = {2018}
}

@article{DBLP:journals/corr/abs-1707-06347,
  author       = {John Schulman and
                  Filip Wolski and
                  Prafulla Dhariwal and
                  Alec Radford and
                  Oleg Klimov},
  title        = {Proximal Policy Optimization Algorithms},
  journal      = {CoRR},
  volume       = {abs/1707.06347},
  year         = {2017},
  url          = {http://arxiv.org/abs/1707.06347},
  eprinttype    = {arXiv},
  eprint       = {1707.06347},
  bibsource    = {dblp computer science bibliography, https://dblp.org}
}

\clearpage

\appendix

\begin{figure*}[t]
    \centering
    \begin{subfigure}[b]{0.23\textwidth}
        \centering
        \includegraphics[width=\linewidth]{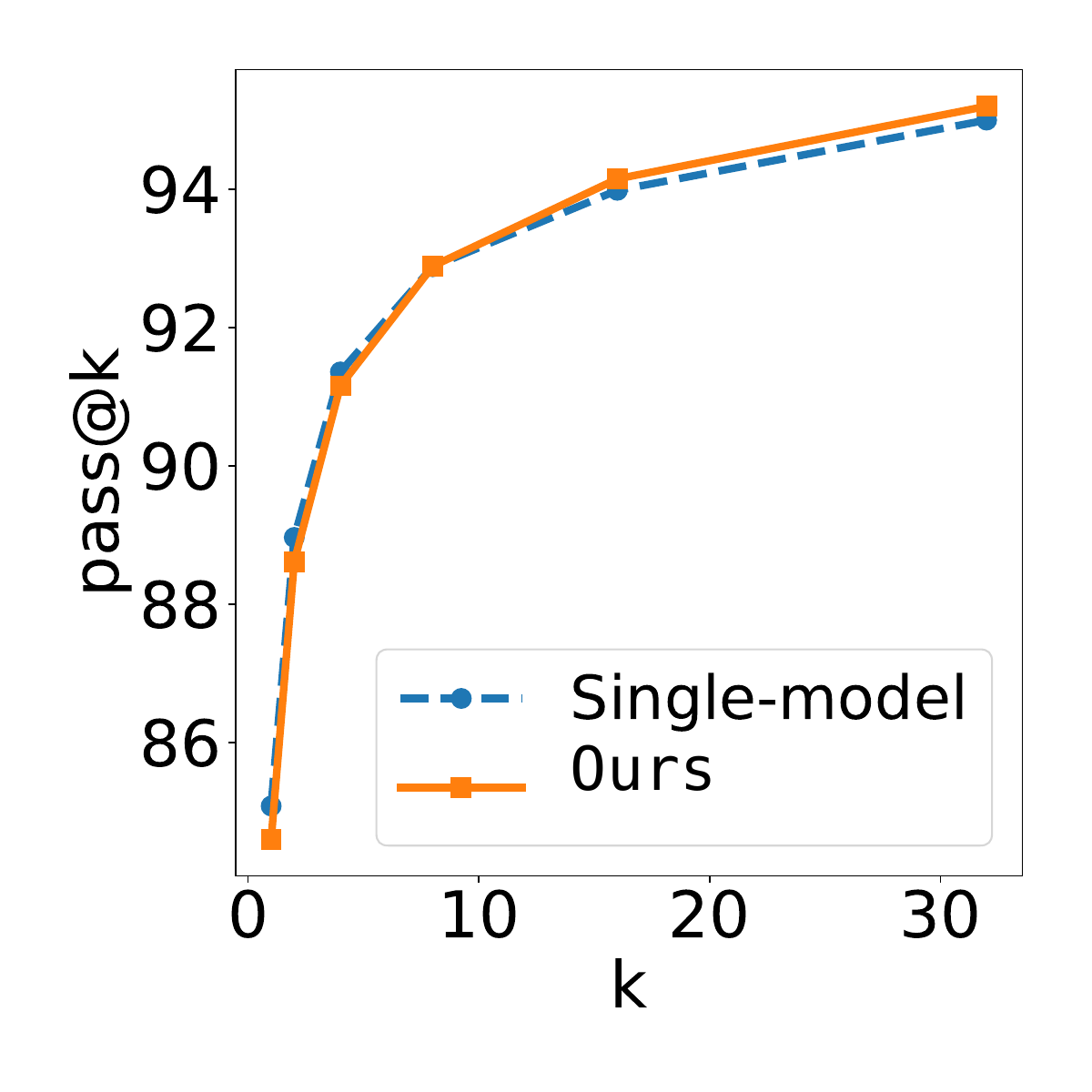}
        \caption{MATH-500}
    \end{subfigure}
    \hfill
    \begin{subfigure}[b]{0.23\textwidth}
        \centering
        \includegraphics[width=\linewidth]{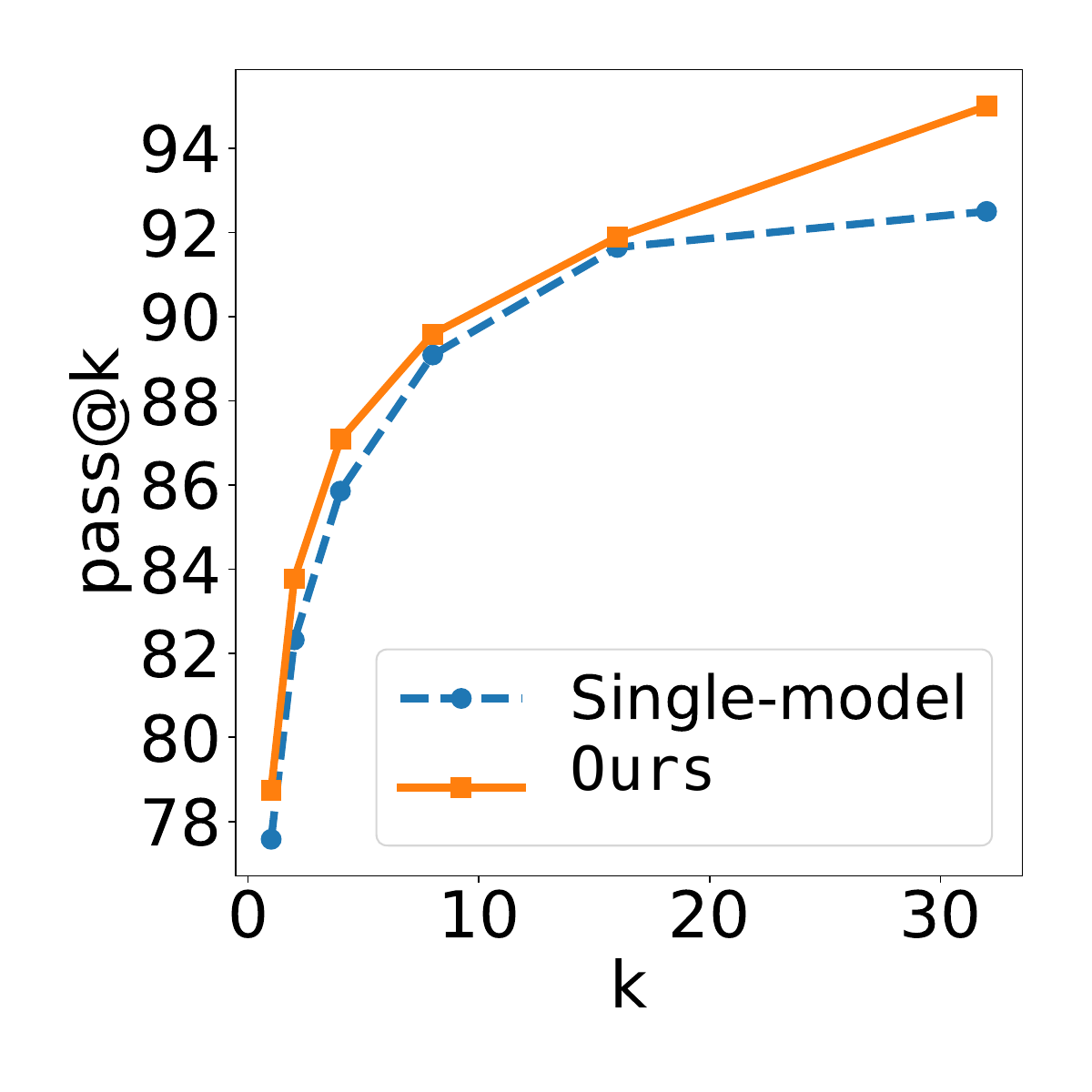}
        \caption{AMC23}
    \end{subfigure}
    \hfill
    \begin{subfigure}[b]{0.23\textwidth}
        \centering
        \includegraphics[width=\linewidth]{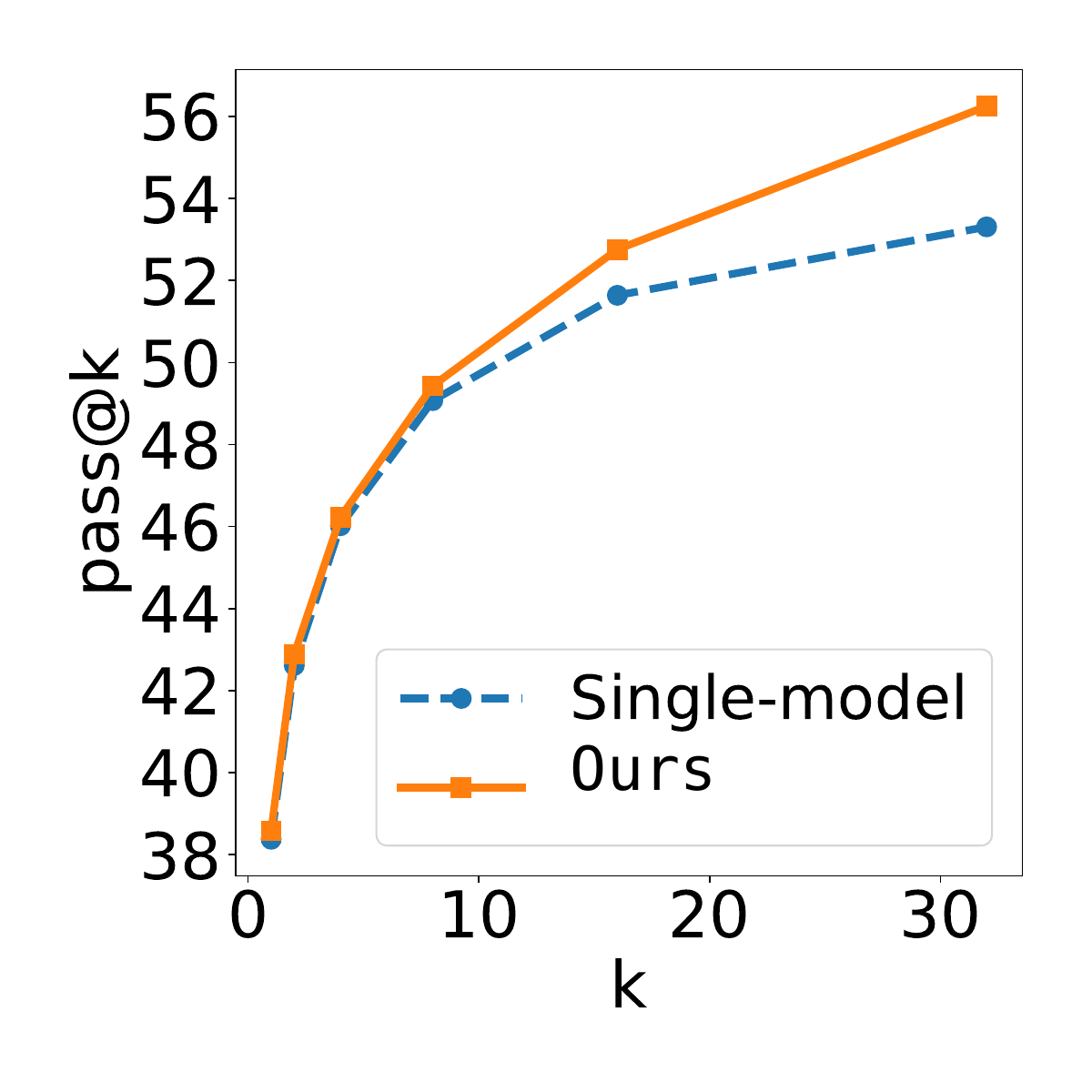}
        \caption{Minerva}
    \end{subfigure}
    \hfill
    \begin{subfigure}[b]{0.23\textwidth}
        \centering
        \includegraphics[width=\linewidth]{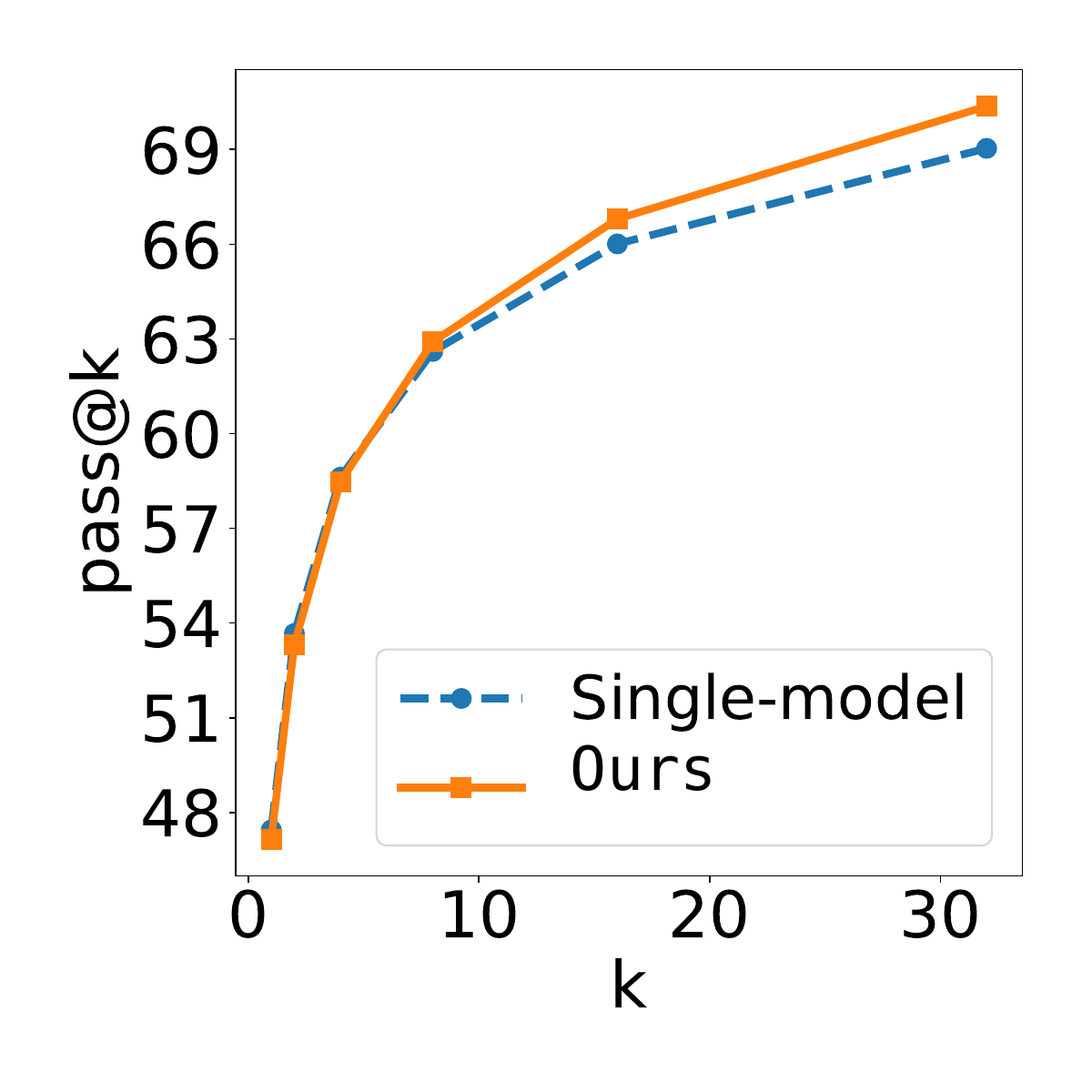}
        \caption{Olympiad}
    \end{subfigure}
    \caption{pass@$k$ of single-model and relay inference.}
    \label{fig:detail_comp}
\end{figure*}

\section{Detailed Comparison Between Single-Model and Relay Inference}
\label{sec:det_comp}

\begin{table*}[t]
\centering
\small
\begin{tabular}{cccccccc}
\toprule
LR & Warmup & Weight Decay & Grad Clip & Clip Low & Clip High & Entropy Coeff & Adv Estimator \\
\midrule
$1e{-}6$ & 10 & 0.1 & 1.0 & 0.2 & 0.28 & 0 & GRPO \\
\bottomrule
\end{tabular}

\vspace{0.5em}
\begin{tabular}{ccccccc}
\toprule
N Responses & Train BSZ & Mini BSZ & Max Prompt Len & Max Resp Len & Loss Agg & Filter Groups \\
\midrule
8 & 32 & 32 & 1024 & 3072 & token-mean & False \\
\bottomrule
\end{tabular}

\vspace{0.5em}
\begin{tabular}{ccccccc}
\toprule
Temperature & Top-p & Top-k & Val Temperature & Val Top-p & Val Top-k & Use Dynamic BSZ \\
\midrule
1.0 & 1.0 & $-1$ & 0.6 & 0.8 & $-1$ & True \\
\bottomrule
\end{tabular}

\vspace{0.5em}
\begin{tabular}{cccccc}
\toprule
KL in Reward & KL Coeff & KL Loss & KL Loss Coeff & Penalty Factor & Metric \\
\midrule
False & 0.0 & False & 0.0 & 1.0 & acc \\
\bottomrule
\end{tabular}
\caption{Hyperparameter settings.}
\label{tab:dapo_hyper}
\end{table*}

Figure~\ref{fig:detail_comp} expands the difference curves of Figure~\ref{fig:single_multi} into the per-benchmark pass@$k$ scores of the two inference strategies.

\section{Proofs}
\label{sec:proofs}

Throughout, a state is $s_t=(q,o_{<t})$ and the transition from $s_t$ under action $o_t$ to $s_{t+1}=(q,o_{\le t})$ is deterministic. A finished response is padded with an end-of-sequence token, after which the state is absorbing and the action is deterministic under every policy, so all sums over $t$ run to the maximum length and $d^{\pi}_t$ is a probability distribution for every $t$. The reward $R(q,o)\in[0,1]$ is received at the end. $\rho_b$ is computed from the cached rollouts of $\pi_b$ and does not depend on ${\boldsymbol{\theta}}$.

\paragraph{Proposition~\ref{prop:exact}.}
For fixed $(q,h)$, $\mathbb{E}_{y\sim\pi(\cdot\mid q,h)}[R]=\sum_y \pi(y\mid q,h)R(q,[h,y])$. Since $\rho_b$ does not depend on ${\boldsymbol{\theta}}$, the gradient passes through the outer expectations and the score-function identity gives
\begin{equation}
\nabla\mathcal{J}_\rho=\mathbb{E}_{q,h,y}\big[R\,\nabla\log\pi(y\mid q,h)\big],
\end{equation}
with $\log\pi(y\mid q,h)=\sum_{t>|h|}\log\pi(o_t\mid s_t)$ for $o=[h,y]$. For a baseline $b(q)$, $\mathbb{E}_{y}[b(q)\nabla\log\pi(y\mid q,h)]=b(q)\nabla\sum_y\pi(y\mid q,h)=0$, so subtracting it leaves the gradient unchanged. On the sample side, each batch is consumed by one update, so the gradient is evaluated at ${\boldsymbol{\theta}}={\boldsymbol{\theta}}_{\text{old}}$, where the continuation-token ratio equals $1$. This lies strictly inside $[1-\varepsilon_{\text{low}},1+\varepsilon_{\text{high}}]$, so both terms of the clipped objective have gradient $\hat A_i\nabla\log\pi(o_{i,t}\mid s_{i,t})$. The group-mean baseline includes the sample itself; algebraically, the centered reward is $(1-\frac{1}{G})$ times the corresponding leave-one-out difference, whose baseline is independent of $y_i$ because the $G$ seeded samples of a prompt are independent (Lemma~\ref{lem:group}). The standard-deviation scaling and token-mean normalization are the same detached sample weights used by on-policy DAPO, but they determine the finite-sample weighting of the estimator. The identity above concerns the continuation policy-gradient component under these fixed sample weights. \hfill$\square$

\paragraph{Proposition~\ref{prop:bias}.}
Write $P^{\mu}_t$ and $P^{\pi}_t$ for the laws of $s_t$, i.e., $d^{\mu}_t$ and $d^{\pi}_t$, and assume $\mu(a\mid s)>0$ wherever $\pi(a\mid s)>0$. Conditioning on $s_t$, the importance ratio makes the inner expectation on-policy:
\begin{align}
&\mathbb{E}_{o_t\sim\mu(\cdot\mid s_t)}\Big[\tfrac{\pi(o_t\mid s_t)}{\mu(o_t\mid s_t)}\Psi_t\nabla\log\pi(o_t\mid s_t)\Big]\nonumber\\
&\quad=\textstyle\sum_a\pi(a\mid s_t)\Psi_t(s_t,a)\nabla\log\pi(a\mid s_t)\nonumber\\
&\quad=g_t(s_t),
\end{align}
so taking $s_t\sim d^{\mu}_t$ gives $\mathbb{E}_{s\sim d^{\mu}_t}[g_t(s)]$, and the same computation with $\mu=\pi$ gives the on-policy gradient. Hence
\begin{align}
&\Big\|\mathbb{E}_{d^{\mu}_t}[g_t]-\mathbb{E}_{d^{\pi}_t}[g_t]\Big\|\nonumber\\
&\quad\le \textstyle\sum_s|d^{\mu}_t(s)-d^{\pi}_t(s)|\,\|g_t(s)\|\nonumber\\
&\quad\le 2G_t\,\mathrm{TV}(d^{\mu}_t,d^{\pi}_t).
\end{align}
Both laws share the prompt distribution, and $s_t$ is generated by $t-1$ successive actions. By the chain rule of the KL divergence, with $\bar K_j$ as in Proposition~\ref{prop:bias},
\begin{equation}
\mathrm{KL}(d^{\mu}_t\,\|\,d^{\pi}_t)=\sum_{j<t}\bar K_j,
\end{equation}
and Pinsker's inequality $\mathrm{TV}\le\sqrt{\mathrm{KL}/2}$ gives $2G_t\,\mathrm{TV}\le G_t(2\sum_{j<t}\bar K_j)^{1/2}$. Finally, $s_t$ is a deterministic function of $s_{t+1}$ (drop the last token), so by the data-processing inequality $\mathrm{TV}(d^{\mu}_t,d^{\pi}_t)\le\mathrm{TV}(d^{\mu}_{t+1},d^{\pi}_{t+1})$. \hfill$\square$

\paragraph{Proposition~\ref{prop:coverage}.}
Let $Q^{\pi}(s,a)$ be the expected reward when action $a$ is taken at $s$ and $\pi$ generates the rest, $V^{\pi}(s)=\mathbb{E}_{a\sim\pi(\cdot\mid s)}Q^{\pi}(s,a)$, and $A^{\pi}=Q^{\pi}-V^{\pi}$. Start from the expression in the proof of Proposition~\ref{prop:exact} and consider one term $t>|h|$. Conditional on $(q,h,o_{\le t})$, the remaining tokens are drawn from $\pi$, so $\mathbb{E}[R\mid q,h,o_{\le t}]=Q^{\pi}(s_t,o_t)$. The right-hand side depends only on $(s_t,o_t)$ because $s_t$ already contains $q$ and $h$; how $s_t$ was reached does not enter. By the tower property, $\mathbb{E}[R\,\nabla\log\pi(o_t\mid s_t)\mathbf{1}\{t>|h|\}]=\mathbb{E}[Q^{\pi}(s_t,o_t)\nabla\log\pi(o_t\mid s_t)\mathbf{1}\{t>|h|\}]$. Subtracting $V^{\pi}(s_t)$, which has zero mean against the score at $s_t$, and summing over $t$ gives the first display, with $d^{\pi}_{\rho,t}$ the law of $s_t$ on the event $\{t>|h|\}$.

In a cycle, batch $1$ is an on-policy batch, whose population continuation gradient is $\nabla\mathcal{J}$ (the case of the empty seed). Batches $2,\dots,T$ are seeded, and by Proposition~\ref{prop:exact} their population continuation gradients are $\nabla\mathcal{J}_\rho$. With ${\boldsymbol{\theta}}$ fixed over the cycle, equal weighting of these population terms gives $\nabla\mathcal{J}+(T-1)\nabla\mathcal{J}_\rho=T\,\nabla\mathcal{J}_T$. The implemented token-mean normalization supplies additional batch-level weights. Applying the first display to the equal-weight population terms, with $d^{\pi}_{\rho,t}=d^{\pi}_t$ for the empty seed, gives the state weighting $d_{T,t}$. \hfill$\square$

\paragraph{The pointwise bound in the remark.}
Since both terms of $d_{T,t}$ are nonnegative, $d_{T,t}\ge\frac{1}{T}d^{\pi}_t$ and $d_{T,t}\ge\frac{T-1}{T}d^{\pi}_{\rho,t}$. Dividing $d^{\pi^\star}_t$ by each lower bound, with $x/0=\infty$ for $x>0$, gives the two terms of the minimum. \hfill$\square$

\paragraph{Lemma~\ref{lem:group}.}
For a prompt $q$, the $G$ behavior rollouts are drawn independently from $\pi_b$ and truncated independently, so the seeds $h_1,\dots,h_G$ are i.i.d.\ from $\rho_b(\cdot\mid q)$, and each continuation $y_i$ is drawn from $\pi(\cdot\mid q,h_i)$ independently of the others. The pairs $(h_i,y_i)$ are therefore i.i.d., and so are the rewards $R_i$, with $\Pr(R_i=1)=\mathbb{E}_{h\sim\rho_b}\Pr_{y\sim\pi}(R=1\mid q,h)=p_\rho$. A group has zero advantage for every rollout exactly when all $G$ rewards are equal, which has probability $p_\rho^G+(1-p_\rho)^G$, and the all-incorrect event has probability $(1-p_\rho)^G$, i.e., $1-\text{pass@}G$ of seeded sampling. For $G\ge2$, $D'(p)=G\big(p^{G-1}-(1-p)^{G-1}\big)$ is negative for $p<\frac12$ and positive for $p>\frac12$. \hfill$\square$

If instead the $G$ continuations of a prompt shared one seed $h$, the rewards would be i.i.d.\ only given $h$, and the probability of a zero-advantage group would be $\mathbb{E}_{h}\big[D(p_h)\big]\ge D(p_\rho)$ by Jensen's inequality, since $D$ is convex and $\mathbb{E}_h p_h=p_\rho$, with $p_h=\Pr_{y\sim\pi}(R{=}1\mid q,h)$. Drawing seeds independently, as Thinking Seeds does, therefore never produces more zero-advantage groups than sharing them at the same marginal success rate.

\section{Detailed Training Parameters}
\label{sec:trainingpara}

Table~\ref{tab:dapo_hyper} lists the basic training settings. For the truncation and refresh parameters we use different configurations across backbones because their context windows differ: $ratio = 50\%$, $k = 32$, $T_{\text{ratio}} = 8$, $T_{\text{entropy}} = 16$ for Qwen2.5-Math-7B, and $ratio = 30\%$, $k = 32$, $T_{\text{ratio}} = T_{\text{entropy}} = 4$ for DeepSeek-R1-Distill-Qwen-1.5B. For the code generation setting with Qwen3-4B-Thinking-2507 we carried the parameters over from the mathematical settings rather than searching again, using $ratio = 30\%$ and $T = 8$.

\section{Per-Benchmark Ablation Results}
\label{sec:ablation_details}

\begin{table*}[t]
    \centering
    \small
    \begin{tabular*}{\textwidth}{@{\extracolsep{\fill}}lccccccc@{}}
    \toprule
        \textbf{Ratio} & \textbf{AIME24} & \textbf{AIME25} & \textbf{MATH-500} & \textbf{AMC23} & \textbf{Minerva} & \textbf{Olympiad} & \textbf{Avg} \\
        \midrule
        0\%  & 34.58 & 17.19 & 83.33 & 71.25 & 37.48 & 45.97 & 48.30 \\
        30\% & 33.65 & 19.38 & \textbf{85.22} & 78.52 & \textbf{38.72} & \textbf{47.90} & 50.56 \\
        50\% & 36.46 & \textbf{19.90} & 85.08 & 77.58 & 38.37 & 47.44 & \textbf{50.81} \\
        70\% & \textbf{37.92} & 18.12 & 84.61 & \textbf{78.91} & 37.32 & 47.50 & 50.73 \\
        90\% & 36.35 & 17.08 & 83.66 & 78.28 & 36.21 & 46.43 & 49.67 \\
        \bottomrule
    \end{tabular*}
    \caption{Truncation ratio sweep ($T=16$, state${+}$action variant, Qwen2.5-Math-7B).}
    \label{tab:ratio_results}

    \vspace{1.0em}
    \begin{tabular*}{\textwidth}{@{\extracolsep{\fill}}lccccccc@{}}
    \toprule
        \textbf{T} & \textbf{AIME24} & \textbf{AIME25} & \textbf{MATH-500} & \textbf{AMC23} & \textbf{Minerva} & \textbf{Olympiad} & \textbf{Avg} \\
        \midrule
        1  & 34.58 & 17.19 & 83.33 & 71.25 & 37.48 & 45.97 & 48.30 \\
        2  & \textbf{37.40} & 17.08 & 84.41 & 79.22 & 38.51 & 47.23 & 50.64 \\
        4  & 34.48 & \textbf{20.52} & 84.46 & 77.34 & 38.15 & 47.74 & 50.45 \\
        8  & 36.35 & 19.90 & 83.25 & \textbf{79.84} & \textbf{39.28} & \textbf{48.32} & \textbf{51.16} \\
        16 & 36.46 & 19.90 & \textbf{85.08} & 77.58 & 38.37 & 47.44 & 50.81 \\
        \bottomrule
    \end{tabular*}
    \caption{Refresh-interval sweep ($ratio=50\%$, state${+}$action variant, Qwen2.5-Math-7B).}
    \label{tab:t_results}
\end{table*}

Tables~\ref{tab:ratio_results} and~\ref{tab:t_results} give the per-benchmark scores summarized by Table~\ref{tab:ablation_avg}, under length-ratio-based truncation. The $ratio=0\%$ and $T=1$ rows are on-policy training.

\section{Training Dynamics}
\label{sec:training_dynamics}

\begin{figure}[t]
    \centering
    \includegraphics[width=0.78\columnwidth]{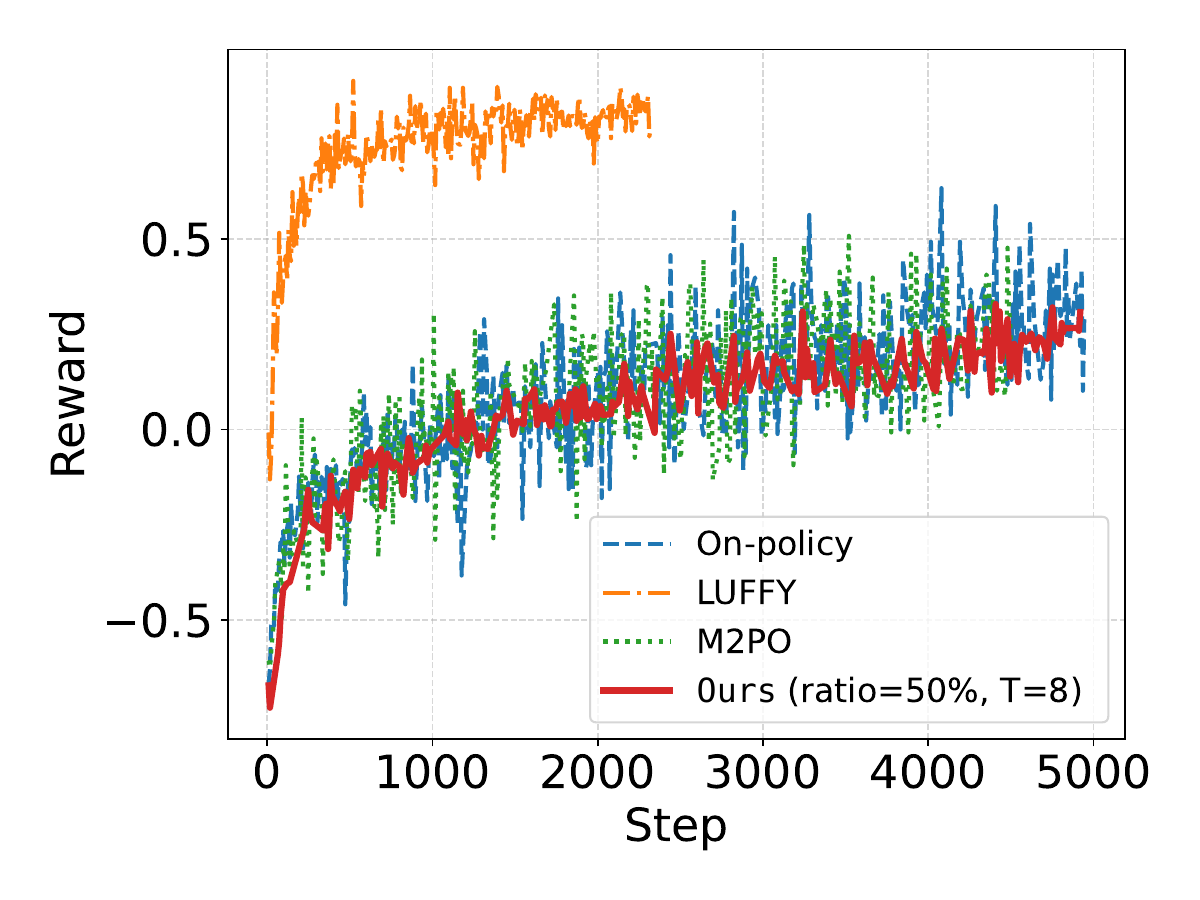}
    \caption{Training reward curves.}
    \label{fig:reward}
\end{figure}

In the plotted run (Figure~\ref{fig:reward}), the Thinking Seeds reward curve has smaller visible fluctuations and a smoother trajectory than the on-policy run. This is a qualitative observation from one run; it is not an estimate of training-seed variance.

\section{Computational Overhead Details}
\label{sec:overhead_details}

In each cycle of $T$ training steps, the current policy, which serves as $\pi_{{\boldsymbol{\theta}}_\text{behavior}}$ for that cycle, generates rollouts for all $T$ batches in a single generation phase and records their token log-probabilities, which later serve as the prefix-side denominators of the importance ratio. The first batch is consumed directly, and only batches $2,\dots,T$ are truncated and completed by the then-current policy. Since the mini-batch size equals the training batch size (Table~\ref{tab:dapo_hyper}), $\pi_{{\boldsymbol{\theta}}_\text{old}}$ is likewise only a cached log-probability tensor. The number of resident models and the number of weight synchronizations to the inference engine are therefore the same as in on-policy training, and the only additional state is a host-side buffer of token ids and log-probabilities, about 100\,MB at $T=16$ in the Qwen2.5-Math-7B setting.

The measured wall-clock time per training step ($1.1$--$1.2\times$ that of on-policy training) is lower than the token factor $1+(1-ratio)\frac{T-1}{T}$ for two reasons. First, pooling $T$ batches into one generation phase pays the long-tail latency of the slowest sequences once rather than $T$ times. Second, completing a truncated rollout only prefills the prefix instead of decoding it.

\end{document}